\documentclass[preprint,12pt]{elsarticle}

\usepackage{amssymb}

\usepackage{lineno}

\usepackage{bm}
\usepackage{amsfonts,amsmath,amssymb}

\usepackage{booktabs}
\usepackage{multirow}
\usepackage{pifont}
\usepackage[table]{xcolor}
\usepackage{subcaption}

\usepackage{lscape}

\usepackage{siunitx}
\usepackage{setspace}
\usepackage{float}

\usepackage[strings]{underscore}

\definecolor{cb-green-sea}  {RGB}{  0, 146, 146}
\definecolor{cb-burgundy}   {RGB}{146,   0,   0}

\newcommand{\eqendp}{\,\text{.}} 

\newcommand{\cmark}{\textcolor{cb-green-sea}{\ding{51}}}
\newcommand{\xmark}{\textcolor{cb-burgundy}{\ding{55}}}

\journal{}

\begin{document}

\begin{frontmatter}



\title{Post-hoc Orthogonalization for Mitigation of Protected Feature Bias in CXR Embeddings}


\affiliation[inst1]{organization={Department of Statistics, LMU Munich},
            addressline={Ludwigstr. 33}, 
            city={Munich},
            postcode={80539}, 
            state={Bavaria},
            country={Germany}}

\affiliation[inst2]{organization={Department of Radiology, University Hospital, LMU Munich},
            addressline={Marchioninistr. 15}, 
            city={Munich},
            postcode={81377}, 
            state={Bavaria},
            country={Germany}}

\affiliation[inst3]{organization={Munich Center for Machine Learning},
            addressline={Geschwister-Scholl-Platz 1}, 
            city={Munich},
            postcode={80539}, 
            state={Bavaria},
            country={Germany}}

\author[inst1,inst2,inst3]{Tobias Weber}
\author[inst2,inst3]{Michael Ingrisch}
\author[inst1,inst3]{Bernd Bischl}
\author[inst1,inst3]{David R\"ugamer}

\begin{abstract}
\textit{Purpose:}
To analyze and remove protected feature effects in chest radiograph embeddings of deep learning models.
\newline
\textit{Methods:}
An orthogonalization is utilized to remove the influence of protected features (e.g., age, sex, race) in CXR embeddings, ensuring feature-independent results. 
To validate the efficacy of the approach, we retrospectively study the MIMIC and CheXpert datasets using three pre-trained models, namely a supervised contrastive, a self-supervised contrastive, and a baseline classifier model.
Our statistical analysis involves comparing the original versus the orthogonalized embeddings by estimating protected feature influences and evaluating the ability to predict race, age, or sex using the two types of embeddings.
\newline
\textit{Results:}
Our experiments reveal a significant influence of protected features on predictions of pathologies.
Applying orthogonalization removes these feature effects.
Apart from removing any influence on pathology classification, while maintaining competitive predictive performance, orthogonalized embeddings further make it infeasible to directly predict protected attributes and mitigate subgroup disparities.
\newline
\textit{Conclusion:}
The presented work demonstrates the successful application and evaluation of the orthogonalization technique in the domain of chest X-ray image classification.
\end{abstract}



\end{frontmatter}


\section{Introduction}

At the interdisciplinary intersection between computer science and radiological imaging, the applications of deep learning for the analysis of chest radiographs (CXR) have garnered considerable attention.
The prominence of deep learning for CXR imaging is largely due to the availability of extensive, annotated datasets, such as \cite{johnson2019mimic, irvin2019chexpert, weber2023cascaded}.
These datasets enable the application of various data-intensive algorithms across diverse research questions. 
Among others, these tasks include classification \cite{irvin2019chexpert}, image generation \cite{weber2023cascaded, chambon2022roentgen}, image manipulation \cite{weber2022implicit}, report generation \cite{liu2019clinically}, and self-supervised pre-training \cite{sellergren2022simplified, cho2023chess}.
With growing public interest and application in practice, there is an escalating demand for fairness, transparency, and explicability of such machine learning models.
This need is paramount in the context of clinical decision-making, where the primary concern is the patient's outcome.

Of particular concern is the potential for bias in predictive models, which might lead to serious consequences in downstream applications.
The work of Gichoya et al. \cite{banerjee2021reading, gichoya2022ai} shows that deep learning models can predict a patient's race in CXRs with high certainty, even when clinical experts cannot.
Building on this observation, the recent study by \cite{glocker2023algorithmic} demonstrates that features like age, sex, and race are already encoded in deep feature representations of neural CXR classifiers and thus can have a strong influence on downstream prediction tasks.
Even more concerning, this property not only affects classifiers but also extends to embeddings from constrastive models, e.g., Google's \textit{Chest X-ray Foundation Model} \cite{sellergren2022simplified, glocker2022risk}.

In the following, we define the term \textit{protected features} as intrinsic characteristics typically protected by law against discrimination and bias.
Not correcting for influences of protected features in predictions and representations used for decision-making can have detrimental effects.
This could lead to CXR classifiers not fully generalizing to certain subgroups as found in \cite{ahluwalia2023subgroup} or to exhibit fairness concerns \cite{seyyed2020chexclusion}.
As \cite{glocker2023algorithmic} point out, ensuring fairness and equal treatment across all subgroups is a major challenge as training datasets tend to be imbalanced for specific populations.

The purpose of this work is twofold:
First, we evaluate whether protected features significantly contribute to model predictions in classifiers based on CXR embeddings using three publicly available models and two open datasets.
Second, we investigate whether a recently proposed orthogonalization procedure \cite{rugamer2023new, rugamer2023semi} can remove the influence of the protected features and lead to unbiased deep representations of CXR images.

\section{Materials and Methods}

\subsection{Orthogonalization}

\begin{figure}[t]
    \centering
    \includegraphics{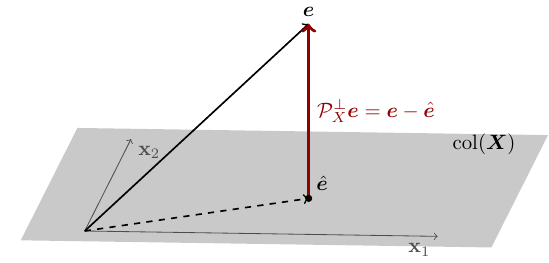}
    \caption{
    Geometric visualization of the orthogonalization method.
    The column space of the protected features ($\text{col}(\bm X)$) contains all the possible vectors that can be formed by taking linear combinations of the respective features, i.e. the hypothesis space of a linear model. For an embedding vector $\bm e \in \bm E$, the orthogonalization is equivalent to the residual between $\bm e$ and its projection onto $\text{col}(\bm X)$. With $\mathcal{P}_X^\bot \bm e$ being perpendicular to $\text{col}(\bm X)$, the influence of protected features in $\bm e$ is neutralized.
}
    \label{fig:ortho-overview}
\end{figure}

In the following, we introduce the orthogonalization more formally. 
Figure~\ref{fig:ortho-overview} shows a graphical illustration of the orthogonalization operation.
Let $\bm E \in \mathbb{R}^{n \times d}$ be the embedding matrix of dimension $d$ for $n$ samples, and $\bm X \in \mathbb{R}^{n \times p}, n \geq p,$ a design matrix containing the information of $p$ protected features for each sample.
Orthogonalization aims to eliminate feature effects of $\bm X$ in $\bm E$, i.e., to remove information about the protected features from the embedding.
This can be achieved by projecting $\bm{E}$ onto the space orthogonal to the space spanned by the columns of $\bm X$.
For this, we construct $\bm{Q}_1 \in \mathbb{R}^{n\times q}$ using the QR-decomposition $\bm{X} = \bm{QR} \in \mathbb{R}^{n\times p}$ with orthogonal matrix $\bm{Q} = [\bm{Q}_1, \bm{Q}_2]$, $\bm{Q}_2 \in \mathbb{R}^{n\times(n-q)}$, and upper-triangular matrix $\bm{R}\in\mathbb{R}^{n\times p}$. $\bm{Q}_1$ corresponds to the non-zero rows in $\bm{R}$ and can be used to define linear projections. The QR decomposition, in this case, is simply an efficient way to compute a projection matrix.
The projection matrix $\mathcal{P}_X$, which projects an object, e.g.~$\bm E$, onto the space spanned by the columns of $\bm{X}$, is then given by $\bm{Q}_1\bm{Q}_1^\top$, and its orthogonal complement $\mathcal{P}_X^\bot = \bm{I} - \mathcal{P}_X$ with $n$-dimensional identity matrix $\bm{I}$ \cite{ben2003generalized}.
Using this orthogonal projection matrix, we can obtain the corrected embedding 
\begin{equation} \label{eq:ortho}
    \widetilde{\bm{E}} = \mathcal{P}_X^\bot  \bm E = (\bm{I} - \bm{Q}_1\bm{Q}_1^\top) \bm E \eqendp
\end{equation}
The projection thereby puts the embedding information into a space with values that any linear combination of the values of protected features cannot represent. The Frisch-Waugh-Lovell theorem \cite{lovell2008simple} then guarantees the subtraction of all $\bm{X}$-influence from $\widetilde{\bm{E}}$.
In particular, this will result in a zero influence of any of the features in $\bm{X}$ on $\bm{E}$ when checking with a significance-test or linear model.

In order to apply this procedure for new unseen data $\bm E^*$, i.e., inference or test data, we derive $\mathcal{P}_{X^*} = \bm{Q}_1^*\bm{Q}_1^{*\top}$ from the associated protected features $\bm X^*$ and subsequently compute the orthogonalized embedding $\widetilde{\bm{E}}^* = (\bm{I} - \mathcal{P}_{X^*}) \bm E^*$.

\subsection{Datasets}

 We retrospectively evaluate the methodology on the MIMIC \cite{johnson2019mimic} and CheXpert \cite{irvin2019chexpert} datasets (cf.~Table~\ref{tab:data} for detailed information and demographics), which both contain patients from US hospitals.
The racial groups are derived in the same fashion as in \cite{gichoya2022ai, glocker2022risk}.
We apply the \textit{U-zeros} strategy from \cite{rajpurkar2017chexnet} to define binary pathology labels, i.e., labels marked as uncertain are considered as negative labels.

For MIMIC, we utilize the recommended training (181,342 scans) and test split (3,041 scans) \cite{johnson2019mimic}, which are contained in the published embeddings\cite{sellergren2023} of the CXR foundation model \cite{sellergren2022simplified}.
Our CheXpert subset consists of the same training (76,205 scans) and test (38,240 scans) split previously employed by \cite{gichoya2022ai, glocker2022risk} in their analyses.
\begin{table}[htbp]
    \centering
    \resizebox{\textwidth}{!}{%
    \begin{tabular}{c|cccccc}
         \toprule
         \multicolumn{7}{c}{\textbf{MIMIC}} \\
         \midrule
         \multicolumn{7}{c}{Training set} \\
         \midrule
         & All & White & Black & Asian & Male & Female \\
         Patients               & 42,148  & 31,936 (75.77\%) & 8,398 (19.93\%) & 1,814 (4.30\%) & 20,123 (47.74\%) & 22,025 (52.26\%) \\
         Scans                  & 181,342 & 140,445 (77.45\%) & 33,906 (18,70\%) & 6,991 (3.86\%) & 97,361 (53.69\%) & 83961 (46.21\%) \\
         Age                    & $62.6 \pm 16.6$ & $63.9 \pm 16.3$ & $57.7 \pm 16.7$ & $62.1 \pm 17.8$ & $62.32 \pm 15.8$ &  $63.0 \pm 17.5$\\    
         \midrule
         \multicolumn{7}{c}{Test set} \\
         \midrule
         & All & White & Black & Asian & Male & Female \\
         Patients               & 257  & 205 (79.77\%) & 45 (17.51\%) & 7 (2.72\%) & 141 (54.86\%) & 116 (45.14\%) \\
         Scans                  & 3,041 & 2,235 (73.50\%) & 676 (22.22\%) & 130 (4.27\%) & 1,658 (54.52\%) & 1,383 (45.48\%) \\
         Age                    & $65.8 \pm 12.1$ & $66.2 \pm 12.3$ & $64.1 \pm 11.9$ & $67.4 \pm 9.5$ & $66.0 \pm 11.6$ &  $65.4 \pm 12.8$\\    
         \toprule
         \multicolumn{7}{c}{\textbf{CheXpert}} \\
         \midrule
         \multicolumn{7}{c}{Training set} \\
         \midrule
         & All & White & Black & Asian & Male & Female \\
         Patients               & 25,730  & 20,034 (77.86\%) & 1,751 (6.81\%)) & 3,945 (15.33\%) & 14,165 (55.05\%) & 11,565 (44.95\%) \\
         Scans                  & 76,205 & 59,238 (77.73\%) & 5,596 (7.34\%) & 11,371 (14.92\%) & 44,774 (58.75\%) & 31,431 (41.25\%) \\
         Age                    & $63.1 \pm 17.4$ & $64.3 \pm 17.2$ & $55.7 \pm 17.4$ & $61.6 \pm 17.4$ & $62.5 \pm 17.0$ &  $63.8 \pm 17.9$\\    
         \midrule
         \multicolumn{7}{c}{Test set} \\
         \midrule
         & All & White & Black & Asian & Male & Female \\
         Patients               & 12,866  & 9,956 (77.38\%) & 879 (6.83\%) & 2,031 (15.79\%) & 7,091 (55.11\%) & 5,775 (44.89\%) \\
         Scans                  & 38,240 & 29,844 (78.04\%) & 2746 (7.18\%) & 5,650 (14.278\%) & 22,265 (58.22\%) & 15,975 (41.78\%) \\
         Age                    & $63.3 \pm 17.2$ & $64.2 \pm 17.1$ & $57.4 \pm 16.3$ & $61.1 \pm 17.6$ & $62.8 \pm 16.4$ &  $63.9 \pm 18.3$\\

    \end{tabular}
    }
    \caption{Statistics of the utilized MIMIC and CheXpert subsets per split and subgroups.}
    \label{tab:data}
\end{table}

\subsection{Analyzed Embeddings}

Our analysis includes the embeddings and feature representations of three publicly available and pre-trained architectures, each compromising different architectures and training procedures. Our code repository is available under:\\\texttt{https://github.com/saiboxx/chexray-ortho}.

\paragraph{CXR Foundation Model (CFM; \cite{sellergren2022simplified})}
This EfficientNet-L2 model \cite{tan2019efficientnet} is trained on over 800,000 CXR images from India and the US using a supervised contrastive procedure \cite{khosla2020supervised}.
The model is only available using a restricted application-based API interface that offers $1376$-dimensional feature embeddings.
Thus, we utilize the publicly available and packaged embeddings of the MIMIC dataset \cite{sellergren2023}.

\paragraph{CheSS \cite{cho2023chess}}
With a ResNet-50 \cite{he2016deep} as a basis, CheSS is a contrastive self-supervised model trained with the MoCo v2 strategy \cite{chen2020improved} on 4.8 million CXRs from South Korea. The model yields $2048$-dimensional feature vectors.

\paragraph{CheXNet Classifier (CLF, \cite{rajpurkar2017chexnet})}
The CLF label represents a Densenet-121 classifier trained on MIMIC and CheXpert, respectively. The models are publicly available and obtained from the \textit{torchxrayvision} library \cite{cohen2022torchxrayvision} yielding $1024$-dimensional last-layer feature vectors.

\subsection{Statistical Analysis}

We analyze i) the influence of protected features on pathology classification, ii) the direct prediction of protected information from embeddings, and iii) downstream task performance, each for the original and orthogonalized embedding, respectively.

\paragraph{Evaluating the Influence of Protected Features}
We employ a linear evaluation protocol, using the embedding as the only feature to predict the class label of interest. 
We predict with a logistic regression model, which is optimized in a gradient-based iterative fashion due to its capability of handling high-dimensionality in the feature embeddings, implemented in PyTorch v2.0 \cite{paszke2017automatic}.
The network is trained for 10 epochs with a learning rate of $0.0001$ and batch size of $256$.
All metrics are evaluated over 10 random initializations.
We do not conduct a hyperparameter search as the goal of this paper is not maximal predictive performance, but to analyze and examine the effects of protected features on predictions.

To assess the impact of protected features on the prediction model, we employ a secondary linear model, referred to as the \emph{evaluation model}, to estimate their influence on the predicted logits generated by the primary \emph{prediction model}.
In practical terms, the \emph{evaluation model} assesses the explainability of the predicted logits given protected features as exposure variables.
From the evaluation model, we obtain regression coefficients for age, race, and sex to quantify their influence and to assess their significance via the provided t-statistics and -tests with $n-p-1$ degrees-of-freedom. 

\paragraph{Extracting Protected Features}
We additionally employ three (generalized) linear models to directly predict the protected information from the embedding.
A linear model is used for the prediction of age, a logistic regression for the prediction of sex, and a multinomial regression to predict the three different race categories defined as described in \cite{glocker2022risk}.
All models are again trained using a gradient-based iterative procedure with a learning rate of $0.0001$ and batch size of $256$ in PyTorch v2.0.

\paragraph{Downstream Prediction Performance}
Lastly, we investigate whether orthogonalization affects the subsequent downstream pathology prediction performance using the previously trained CXR prediction models.

\section{Results}

\subsection{Influence of Protected Features on Predictions}\label{sec:influence}

The estimated influence of protected features for the exemplary labels \textit{Pleural Effusion}, \textit{Cardiomegaly}, and \textit{No Finding} is given in Table~\ref{tab:pred-coefs}.
Extended results including the distribution of coefficients and p-values over multiple runs are found in Supplement~\ref{app:influence}.

\paragraph{Without Orthogonalization}
Our results highlight that protected features have a substantial influence on the model outputs, indicated by the small p-values for the majority of coefficients.
In our feature encoding age is divided by 100, thus an increase in one year is equivalent to 0.01.
For example, in the case of the MIMIC CFM embeddings and \textit{Pleural Effusion}, an increase of patient age of ten years results in a significant increase of $0.1 \cdot 4.246 = 0.42$ in the predicted logits, or equivalent, a multiplicative increase in odds of almost 52\% for the prediction model.
\begin{table}[t]
    \centering
    \resizebox{0.95\textwidth}{!}{%
    \begin{tabular}{llccccc}
    \toprule
       &  \textbf{Pathology} & \textbf{Orthogonalized?} & \textbf{Age} &\textbf{Sex[Female]} & \textbf{Race[Black]} & \textbf{Race[Asian]} \\ 
    \midrule
    \multirow{6}{*}{\rotatebox[origin=c]{90}{\textbf{MIMIC CFM}}}
    & \multirow{2}{*}{Pleural Effusion} & \xmark & $\cellcolor{cb-burgundy!15} 4.246$ ($<\num{2e-16}$)& $\cellcolor{cb-burgundy!15} -0.143$ ($<\num{2e-16}$) &\cellcolor{cb-burgundy!15}   $-0.694$ ($<\num{2e-16}$) & $0.029$ ($0.386$) \\
    &                                   & \cmark & $0$ ($1$) & $0$ ($1$) & $0$ ($1$) & $0$ ($1$)\\
    \cmidrule{2-7}
    & \multirow{2}{*}{Cardiomegaly} & \xmark & \cellcolor{cb-burgundy!15} $3.490$ ($<\num{2e-16}$)& \cellcolor{cb-burgundy!15} $-0.078$ ($<\num{2e-16}$) & \cellcolor{cb-burgundy!15}  $0.145$ ($<\num{2e-16}$) & $-0.017$ ($0.428$) \\
    &                                   & \cmark & $0$ ($1$) & $0$ ($1$) & $0$ ($1$) & $0$ ($1$)\\
    \cmidrule{2-7}
    & \multirow{2}{*}{No Finding} & \xmark & \cellcolor{cb-burgundy!15}  $-3.438$ ($<\num{2e-16}$)& \cellcolor{cb-burgundy!15}  $0.216$ ($<\num{2e-16}$) & \cellcolor{cb-burgundy!15}  $0.307$ ($<\num{2e-16}$) & $0.157$ ($0.350$) \\
    &                                   & \cmark & $0$ ($1$) & $0$ ($1$) & $0$ ($1$) & $0$ ($1$)\\
    \bottomrule
    \multirow{6}{*}{\rotatebox[origin=c]{90}{\textbf{MIMIC CheSS}}}
    & \multirow{2}{*}{Pleural Effusion} & \xmark & \cellcolor{cb-burgundy!15} $3.529$ ($<\num{2e-16}$)& $-0.017$ ($0.143$) & \cellcolor{cb-burgundy!15}   $-0.277$ ($<\num{2e-16}$) & \cellcolor{cb-burgundy!15}  $0.070$ ($\num{5e-4}$) \\
    &                                   & \cmark & $0$ ($1$) & $0$ ($1$) & $0$ ($1$) & $0$ ($1$)\\
    \cmidrule{2-7}
    & \multirow{2}{*}{Cardiomegaly} & \xmark & \cellcolor{cb-burgundy!15} $2.658$ ($<\num{2e-16}$)& $-0.007$ ($0.321$) & \cellcolor{cb-burgundy!15}  $0.088$ ($<\num{2e-16}$) & $0.013$ ($0.386$) \\
    &                                   & \cmark & $0$ ($1$) & $0$ ($1$) & $0$ ($1$) & $0$ ($1$)\\
    \cmidrule{2-7}
    & \multirow{2}{*}{No Finding} & \xmark & \cellcolor{cb-burgundy!15}  $-2.974$ ($<\num{2e-16}$)& \cellcolor{cb-burgundy!15} $0.250$ ($<\num{2e-16}$) & \cellcolor{cb-burgundy!15} $0.221$ ($<\num{2e-16}$) & $-0.024$ ($0.150$) \\
    &                                   & \cmark & $0$ ($1$) & $0$ ($1$) & $0$ ($1$) & $0$ ($1$)\\
    \bottomrule
    \multirow{6}{*}{\rotatebox[origin=c]{90}{\textbf{MIMIC CLF}}}
    & \multirow{2}{*}{Pleural Effusion} & \xmark & \cellcolor{cb-burgundy!15} $3.582$ ($<\num{2e-16}$)& \cellcolor{cb-burgundy!15} $-0.193$ ($<\num{2e-16}$) &  \cellcolor{cb-burgundy!15} $-0.275$ ($<\num{2e-16}$) & \cellcolor{cb-burgundy!15} $0.069$ ($\num{0.001}$) \\
    &                                   & \cmark & $0$ ($1$) & $0$ ($1$) & $0$ ($1$) & $0$ ($1$)\\
    \cmidrule{2-7}
    & \multirow{2}{*}{Cardiomegaly} & \xmark & \cellcolor{cb-burgundy!15} $2.992$ ($<\num{2e-16}$)& \cellcolor{cb-burgundy!15} $-0.099$ ($<\num{2e-16}$) &  \cellcolor{cb-burgundy!15} $0.136$ ($<\num{2e-16}$) & $-0.018$ ($0.255$) \\
    &                                   & \cmark & $0$ ($1$) & $0$ ($1$) & $0$ ($1$) & $0$ ($1$)\\
    \cmidrule{2-7}
    & \multirow{2}{*}{No Finding} & \xmark & $ \cellcolor{cb-burgundy!15} -2.907$ ($<\num{2e-16}$)& \cellcolor{cb-burgundy!15} $0.217$ ($<\num{2e-16}$) &  \cellcolor{cb-burgundy!15} $0.207$ ($<\num{2e-16}$) & $-0.018$ ($0.279$) \\
    &                                   & \cmark & $0$ ($1$) & $0$ ($1$) & $0$ ($1$) & $0$ ($1$)\\

    \bottomrule
    \multirow{6}{*}{\rotatebox[origin=c]{90}{\textbf{CheXpert CheSS}}}
    & \multirow{2}{*}{Pleural Effusion} & \xmark & \cellcolor{cb-burgundy!15} $1.679$ ($<\num{2e-16}$)& \cellcolor{cb-burgundy!15} $0.135$ ($<\num{2e-16}$) &  \cellcolor{cb-burgundy!15} $-0.072$ ($\num{2e-5}$) & \cellcolor{cb-burgundy!15} $0.051$ ($\num{1e-4}$) \\
    &                                   & \cmark & $0$ ($1$) & $0$ ($1$) & $0$ ($1$) & $0$ ($1$)\\
    \cmidrule{2-7}
    & \multirow{2}{*}{Cardiomegaly} & \xmark & \cellcolor{cb-burgundy!15} $1.278$ ($<\num{2e-16}$)& $0.004$ ($0.511$) &  \cellcolor{cb-burgundy!15} $0.362$ ($<\num{2e-16}$) & \cellcolor{cb-burgundy!15} $0.112$ ($<\num{2e-16}$) \\
    &                                   & \cmark & $0$ ($1$) & $0$ ($1$) & $0$ ($1$) & $0$ ($1$)\\
    \cmidrule{2-7}
    & \multirow{2}{*}{No Finding} & \xmark & \cellcolor{cb-burgundy!15} $-2.347$ ($<\num{2e-16}$)& \cellcolor{cb-burgundy!15} $0.092$ ($<\num{2e-16}$) &  $-0.027$ ($0.120$) & \cellcolor{cb-burgundy!15} $-0.045$ ($0.001$) \\
    &                                   & \cmark & $0$ ($1$) & $0$ ($1$) & $0$ ($1$) & $0$ ($1$)\\

    \bottomrule
    \multirow{6}{*}{\rotatebox[origin=c]{90}{\textbf{CheXpert CLF}}}
    & \multirow{2}{*}{Pleural Effusion} & \xmark & \cellcolor{cb-burgundy!15} $1.590$ ($<\num{2e-16}$)& $0.025$ ($0.060$) & \cellcolor{cb-burgundy!15}  $-0.067$ ($\num{5e-4}$) & \cellcolor{cb-burgundy!15}  $0.036$ ($\num{0.012}$) \\
    &                                   & \cmark & $0$ ($1$) & $0$ ($1$) & $0$ ($1$) & $0$ ($1$)\\
    \cmidrule{2-7}
    & \multirow{2}{*}{Cardiomegaly} & \xmark & \cellcolor{cb-burgundy!15}  $1.613$ ($<\num{2e-16}$)& \cellcolor{cb-burgundy!15} $-0.090$ ($\num{2e-16}$) &  \cellcolor{cb-burgundy!15} $0.454$ ($<\num{2e-16}$) & \cellcolor{cb-burgundy!15} $-0.081$ ($\num{2e-9}$) \\
    &                                   & \cmark & $0$ ($1$) & $0$ ($1$) & $0$ ($1$) & $0$ ($1$)\\
    \cmidrule{2-7}
    & \multirow{2}{*}{No Finding} & \xmark & \cellcolor{cb-burgundy!15} $-2.356$ ($<\num{2e-16}$)& \cellcolor{cb-burgundy!15} $0.039$ ($0.002$) &  $-0.037$ ($0.057$) & \cellcolor{cb-burgundy!15} $-0.071$ ($<\num{8e-7}$) \\
    &                                   & \cmark & $0$ ($1$) & $0$ ($1$) & $0$ ($1$) & $0$ ($1$)\\

    \bottomrule
    \end{tabular}
    }
    \caption{The average coefficients (with p-values in brackets) for the impact of protected features on predictions for \textit{Pleural Effusion}, \textit{Cardiomegaly}, and \textit{No Finding}. For predicting a scan's race, the prevalent label \textit{White} is chosen as the reference level. Cells containing p-values $<$ 0.05 are colored red marking significant influence. After orthogonalization, all coefficients will be 0 with p-values of 1, as the influence of the orthogonalized features is completely removed from the target.}
    \label{tab:pred-coefs}
\end{table}

Moreover, race also shows a significant influence: being a Black patient instead of a White patient results in a reduction of the predicted probability for \textit{Pleural Effusion}.
The respective estimated additive effect $-0.694$ results in logit values that imply a reduction of 50\% in the odds for the prediction model.
While being Black in the MIMIC CFM embedding is associated with a risk-reducing factor for \textit{Pleural Effusion}, it increases for \textit{Cardiomegaly} with a coefficient of $0.145$.

Being female is generally associated with a negative coefficient with the exception of \textit{Pleural Effusion} in the CheXpert CheSS embeddings ($0.135$).
The coefficients for Asians exhibit relatively small values and, in many cases, p-values greater than 0.05, indicating a lack of statistical significance.
The magnitudes of the coefficients for the label \textit{No Finding}, which indicates the presence of a patient with no abnormalities, are similar to \textit{Pleural Effusion} and \textit{Cardiomegaly}, but with opposite signs.

Overall, the analysis of MIMIC results in higher coefficients than CheXpert.
When evaluating the CheSS embedding, e.g., the age coefficient for \textit{Pleural Effusion} is $3.529$ versus $1.679$.
Other pathologies and protected features, except for the coefficient of the attribute Black, which increases from $0.088$ to $0.362$ between the two datasets.

In summary, the obtained coefficients express and state the existence of biases in downstream predictions.

\begin{figure}[p]
    \centering
    \resizebox{\textwidth}{!}{%
    \begin{tabular}{ccc|cc}
        & \multicolumn{2}{c}{\textbf{PCA components by sex}} & \multicolumn{2}{c}{\textbf{PCA components by} \textit{Pleural Effusion}} \\
        & \textbf{Original} & \textbf{Orthogonalized} & \textbf{Original} & \textbf{Orthogonalized} \\
        \rotatebox[origin=l]{90}{\textbf{MIMIC CFM}} &
        \includegraphics[width=.20\textwidth]{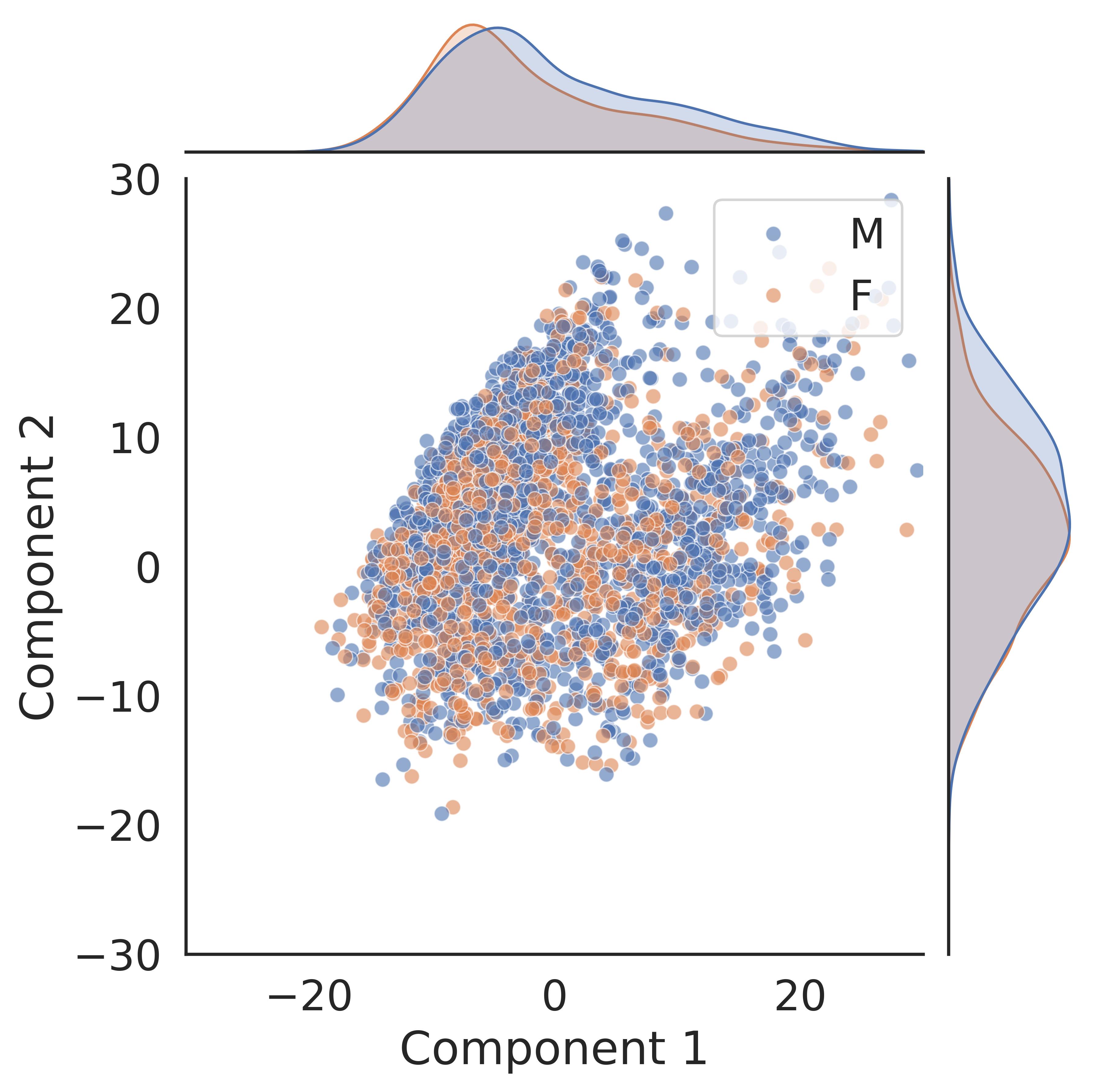} &
        \includegraphics[width=.20\textwidth]{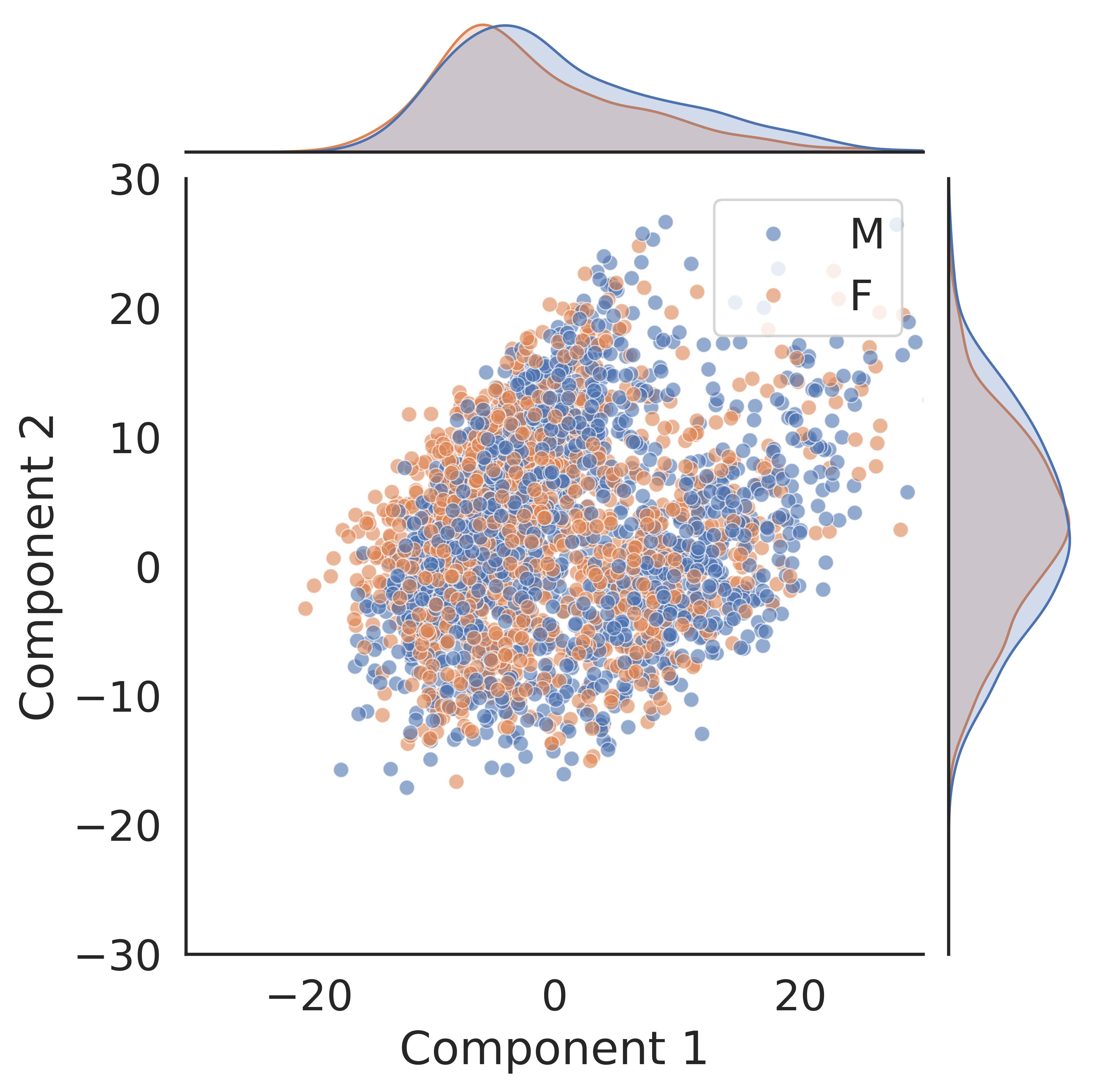} &
        \includegraphics[width=.20\textwidth]{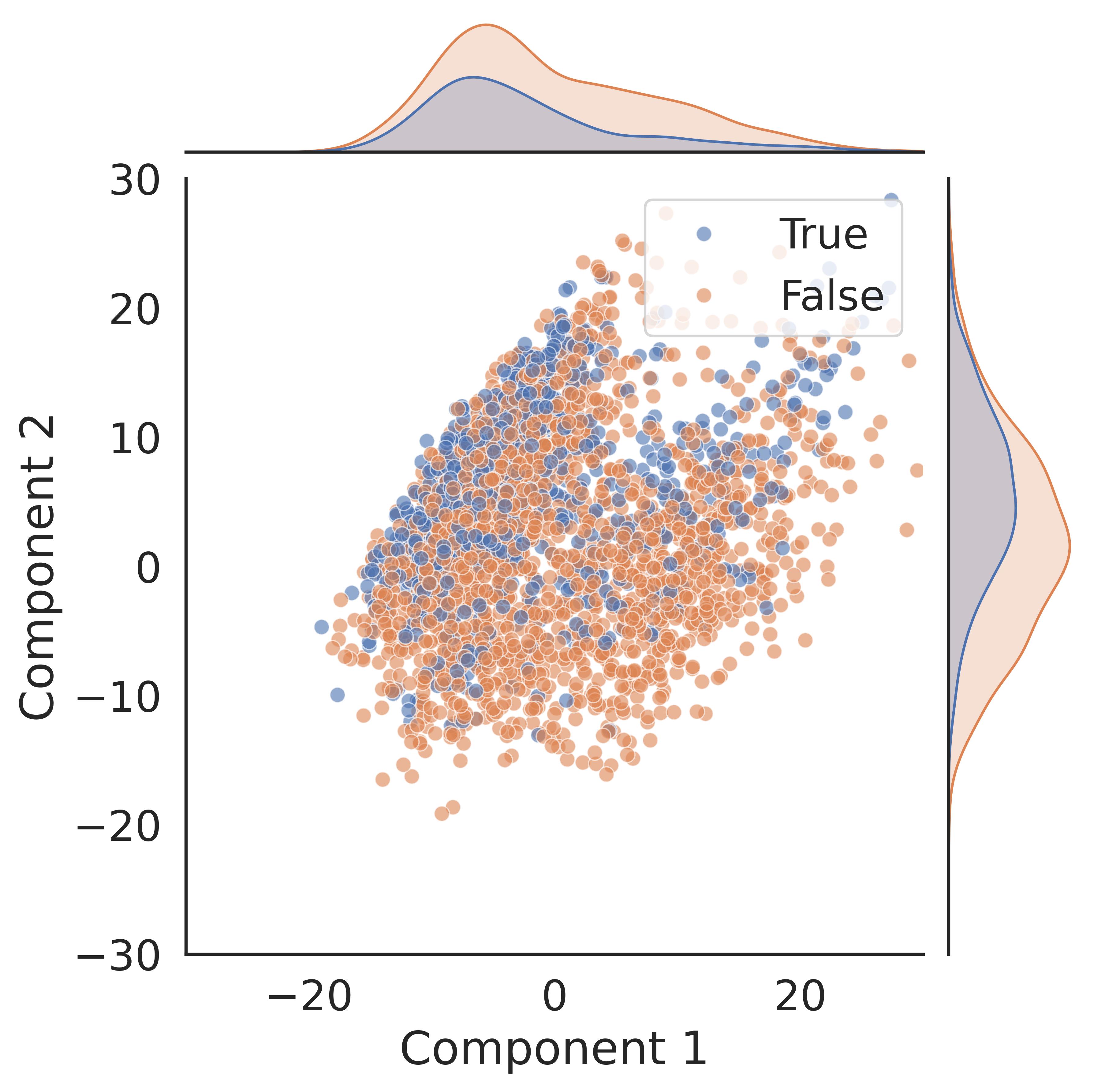} &
        \includegraphics[width=.20\textwidth]{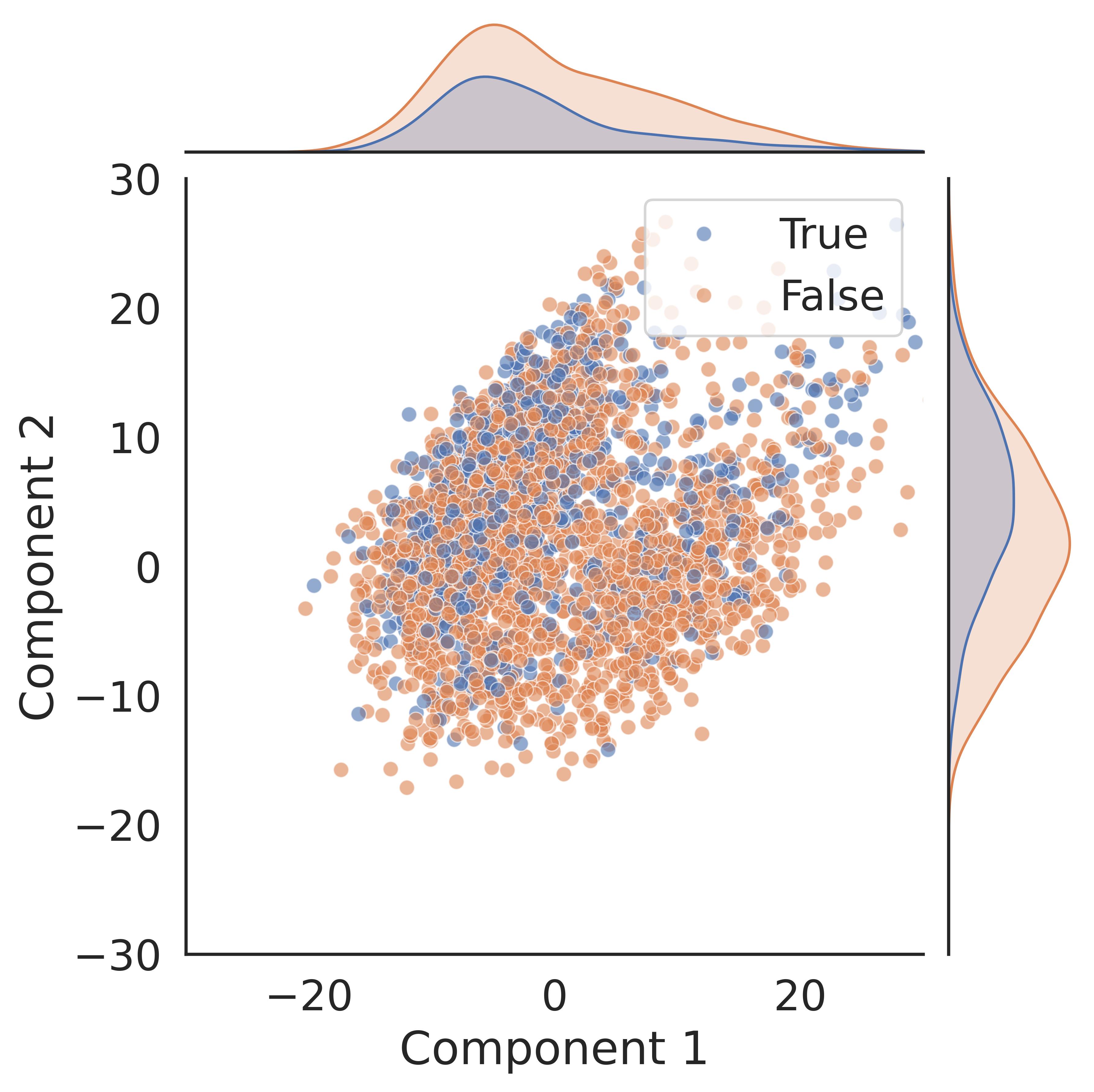} \\
        \midrule
        \rotatebox[origin=l]{90}{\textbf{MIMIC CheSS}} &
        \includegraphics[width=.20\textwidth]{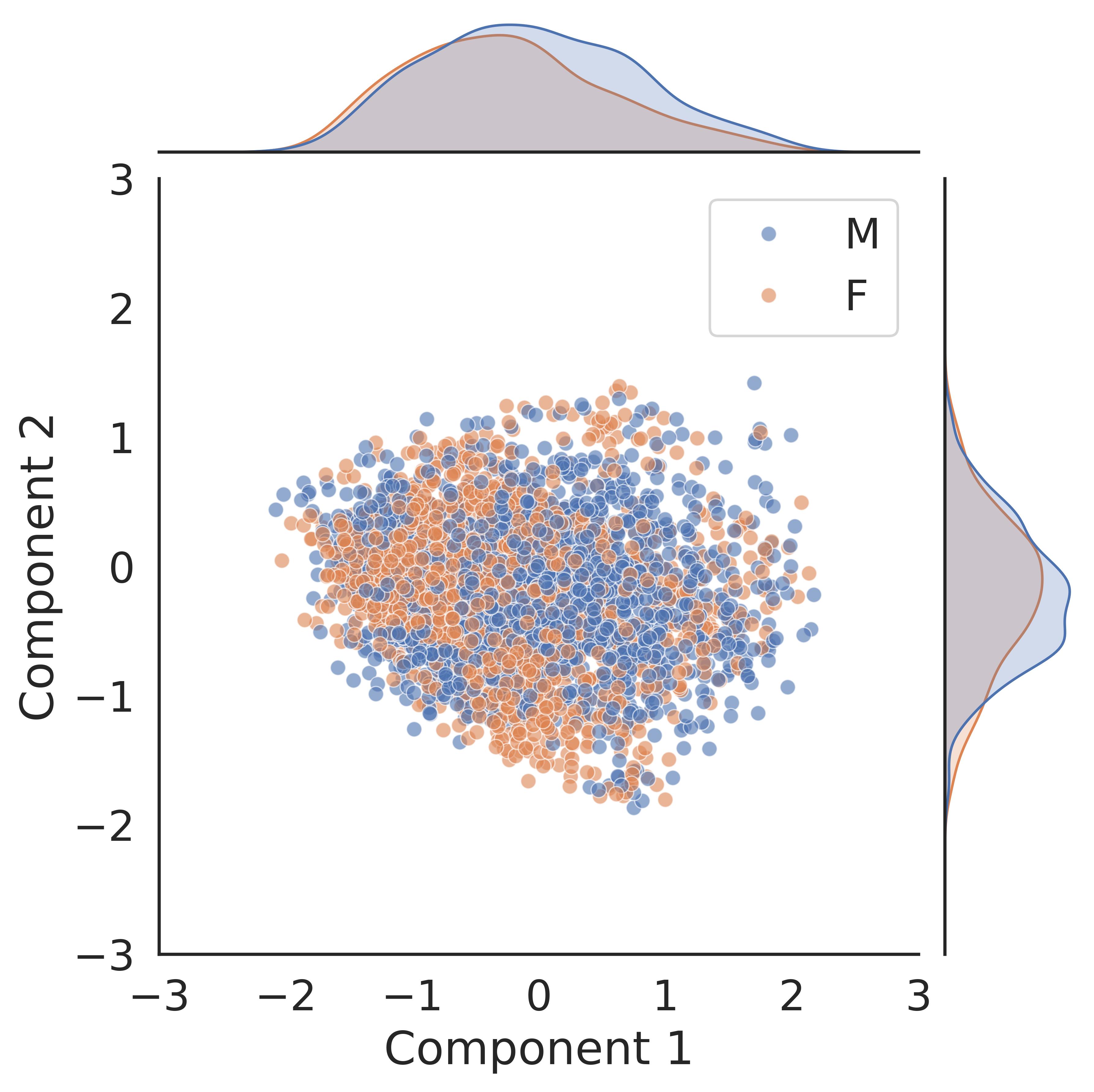} &
        \includegraphics[width=.20\textwidth]{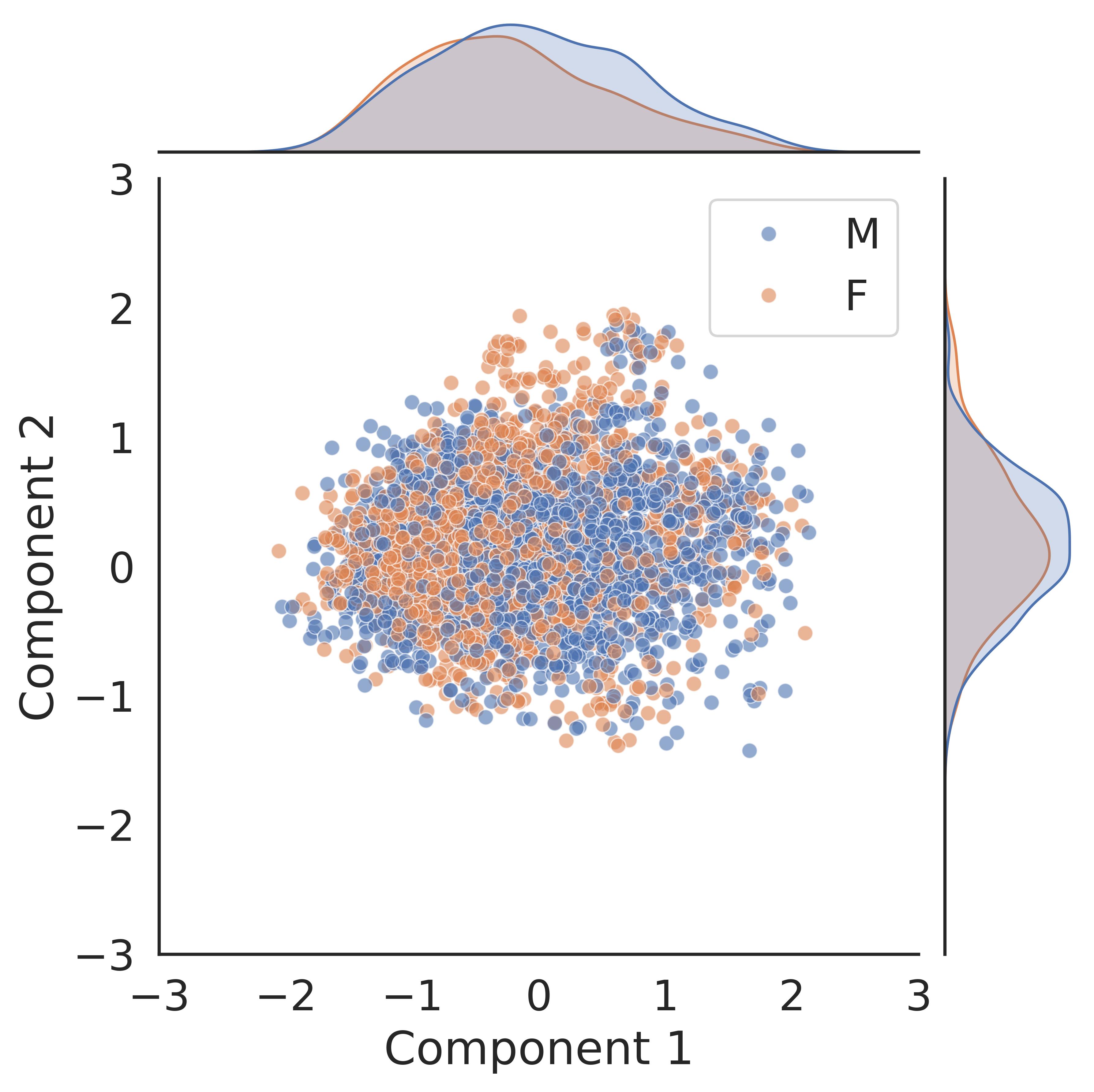} &
        \includegraphics[width=.20\textwidth]{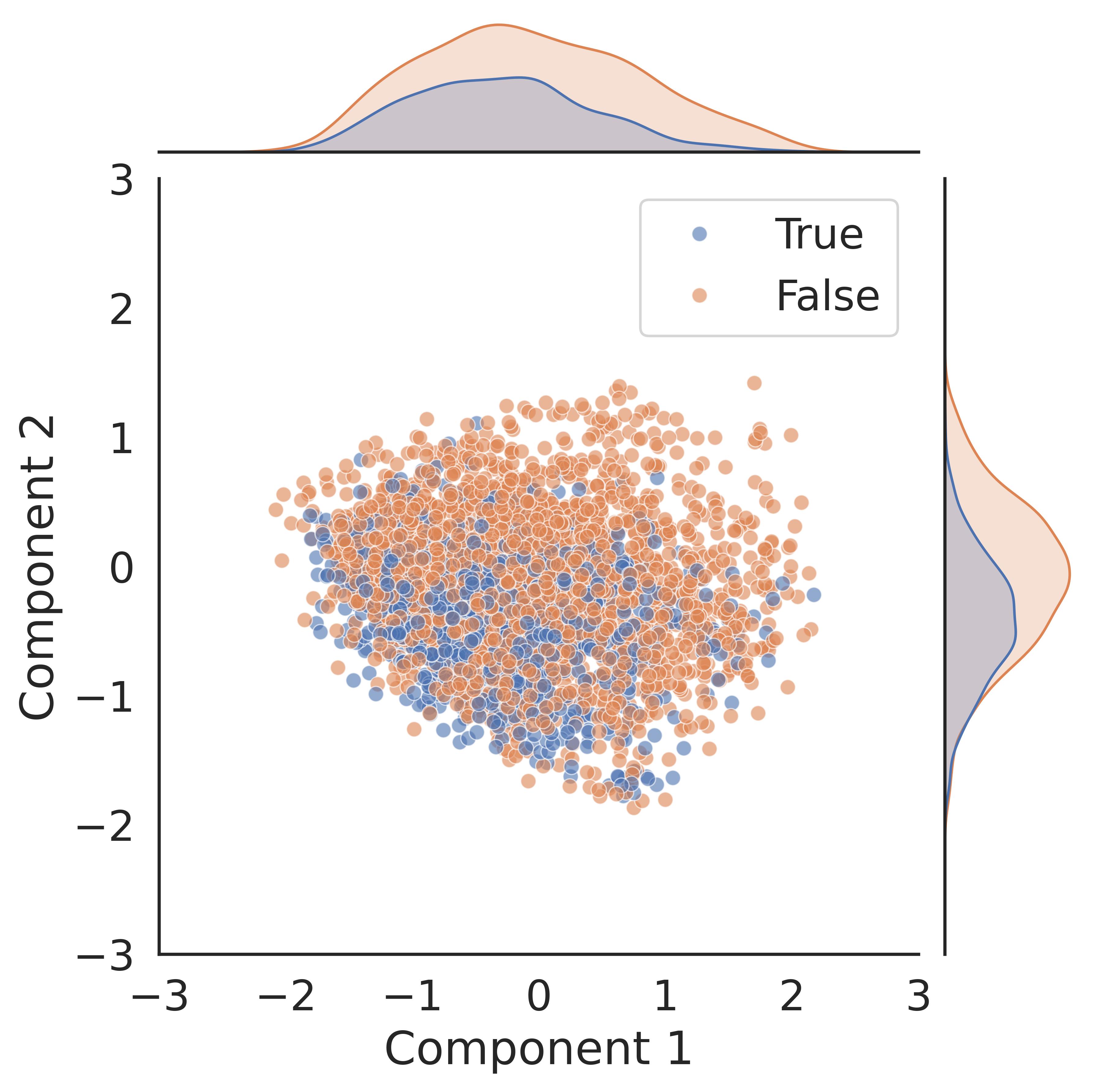} &
        \includegraphics[width=.20\textwidth]{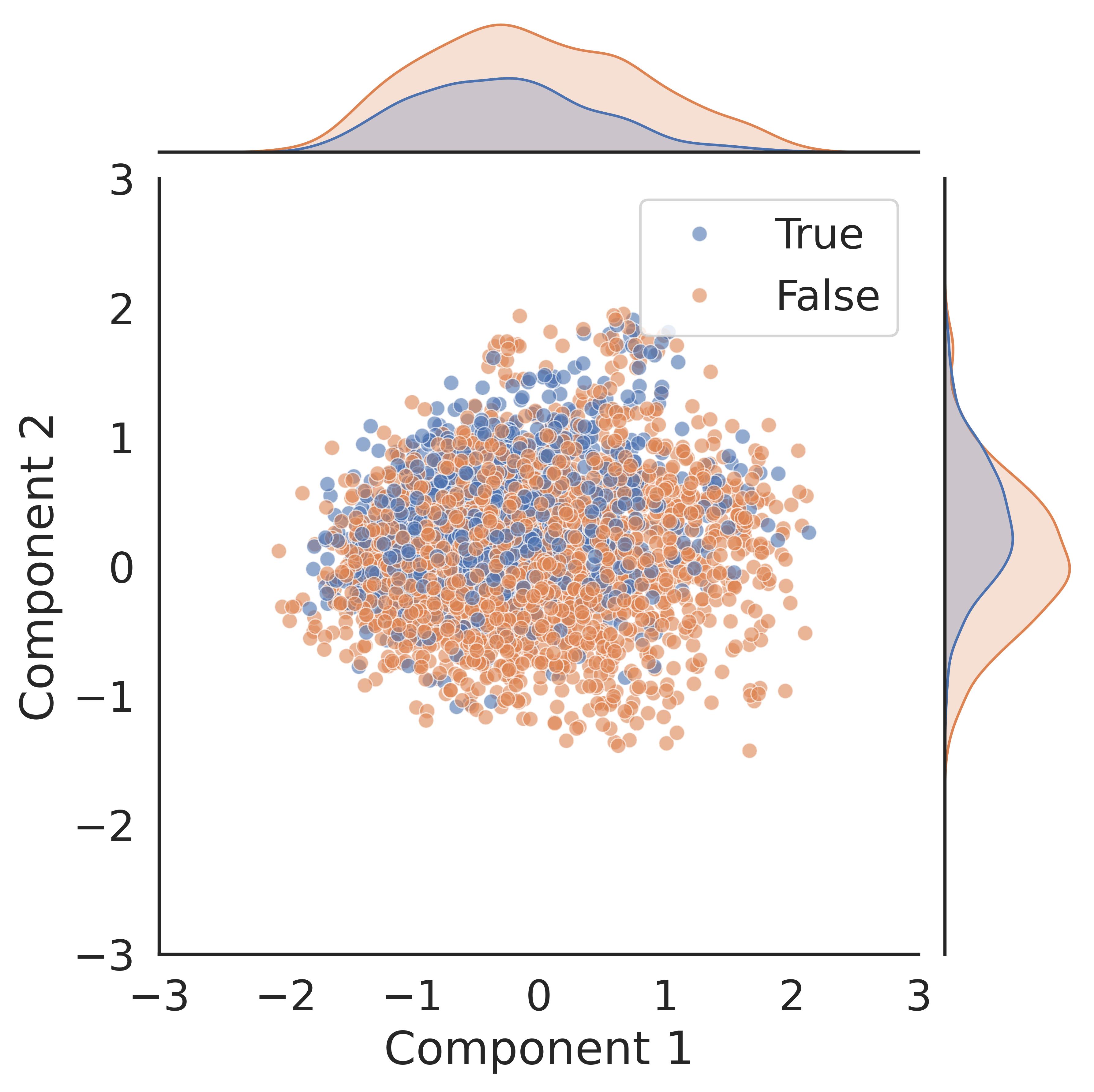} \\
        \midrule
        \rotatebox[origin=l]{90}{\textbf{MIMIC CLF}} &
        \includegraphics[width=.20\textwidth]{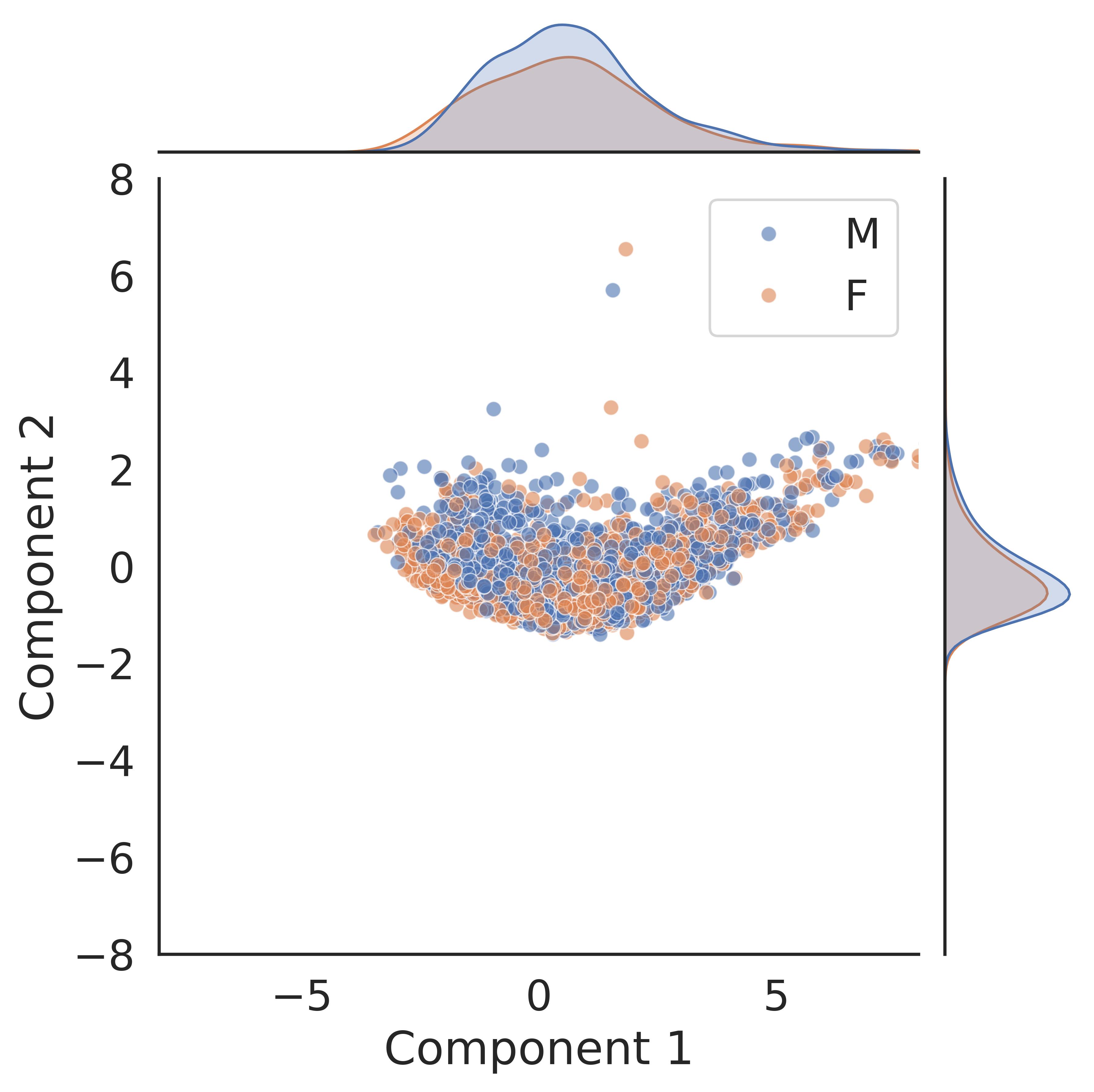} &
        \includegraphics[width=.20\textwidth]{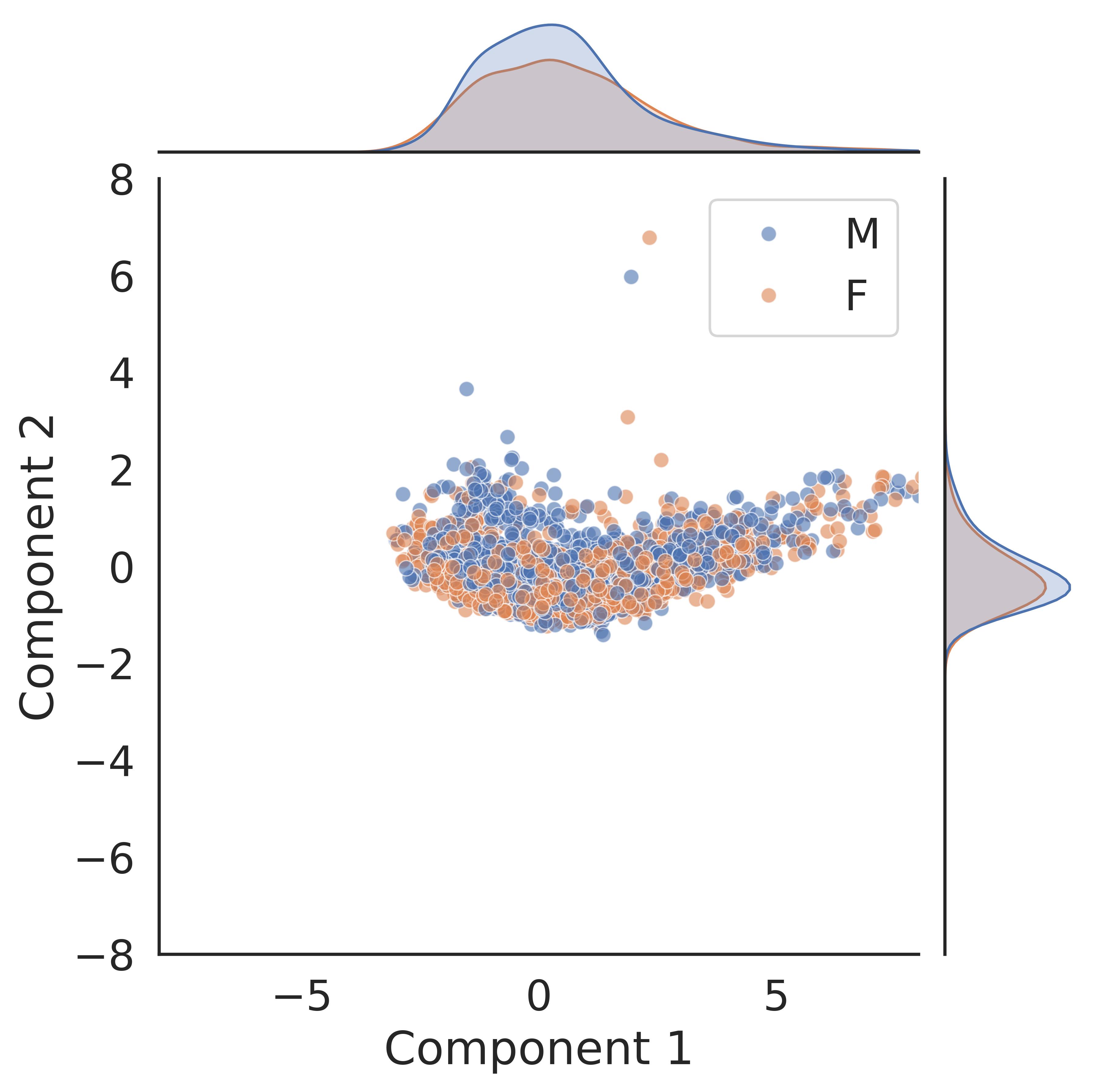} &
        \includegraphics[width=.20\textwidth]{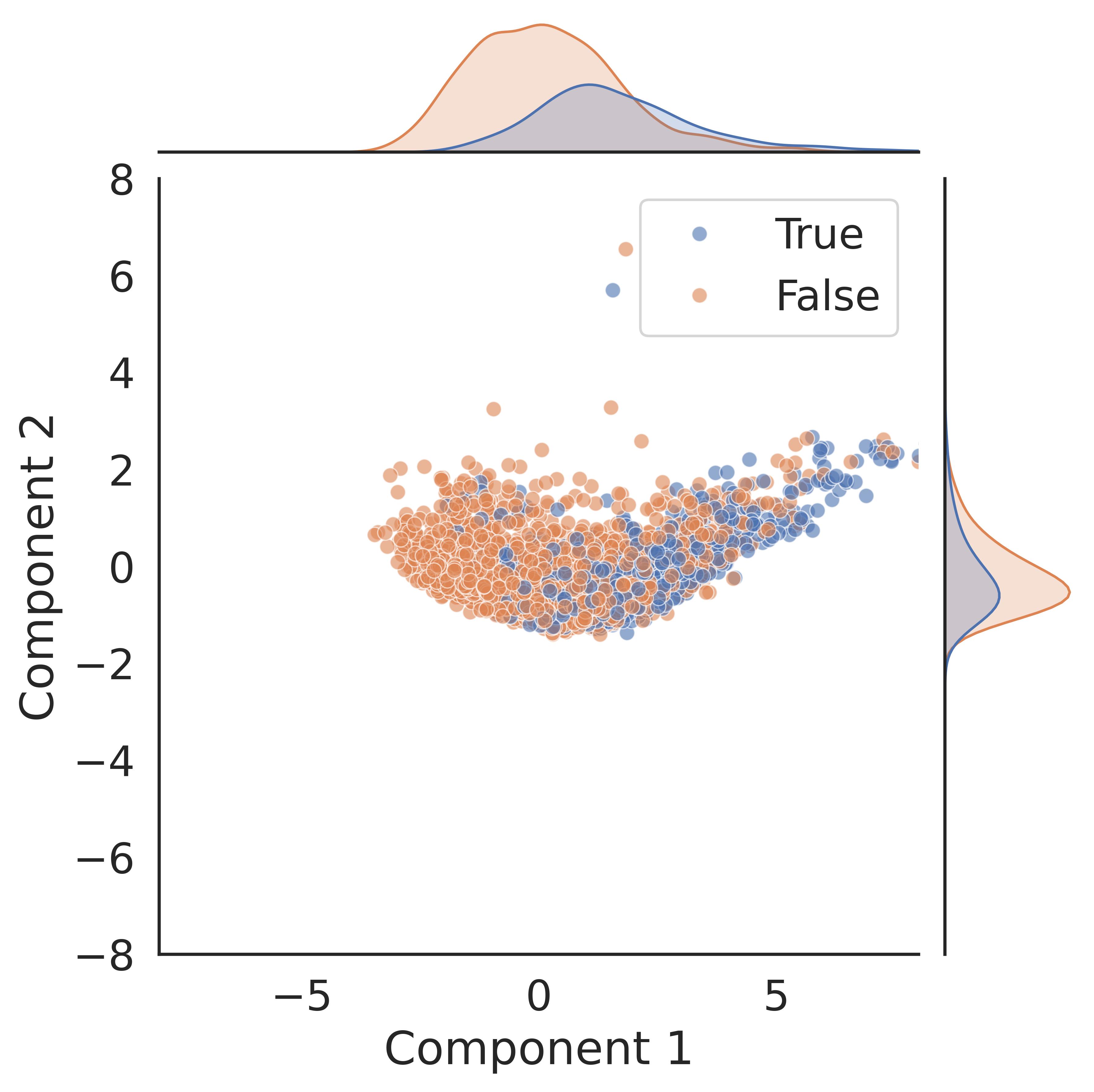} &
        \includegraphics[width=.20\textwidth]{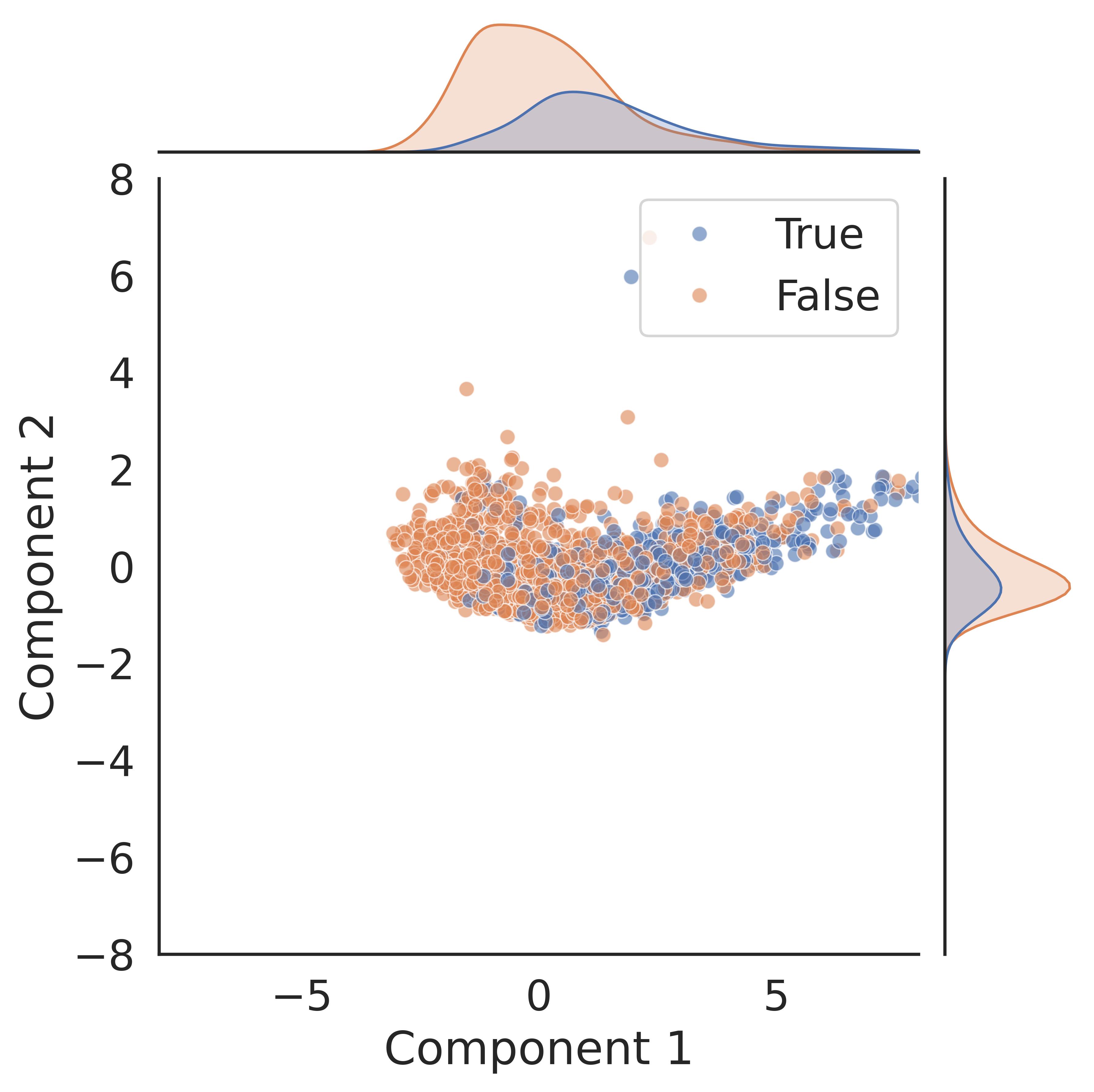} \\
        \midrule
        \rotatebox[origin=l]{90}{\textbf{CheXpert CheSS}} &
        \includegraphics[width=.20\textwidth]{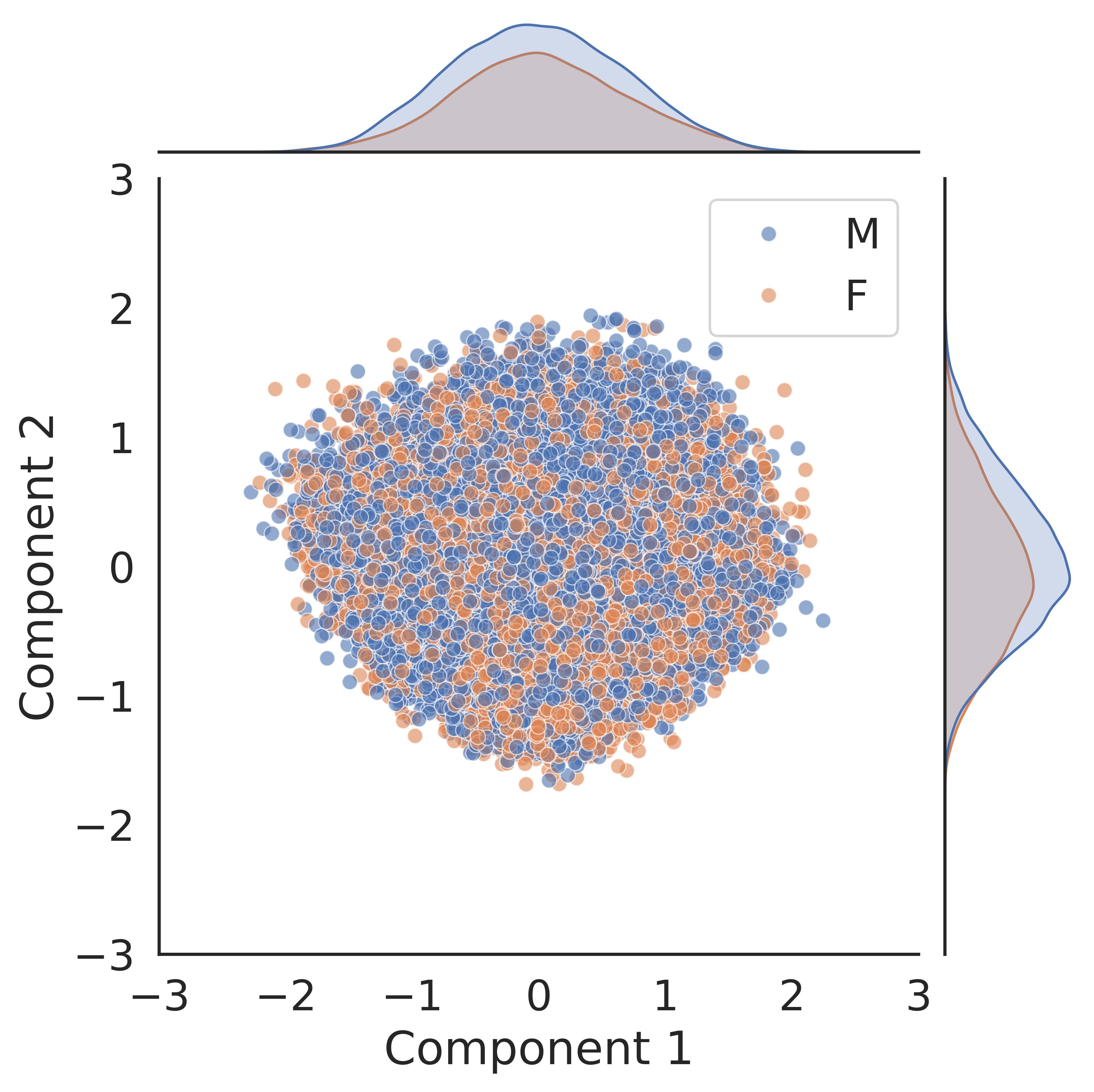} &
        \includegraphics[width=.20\textwidth]{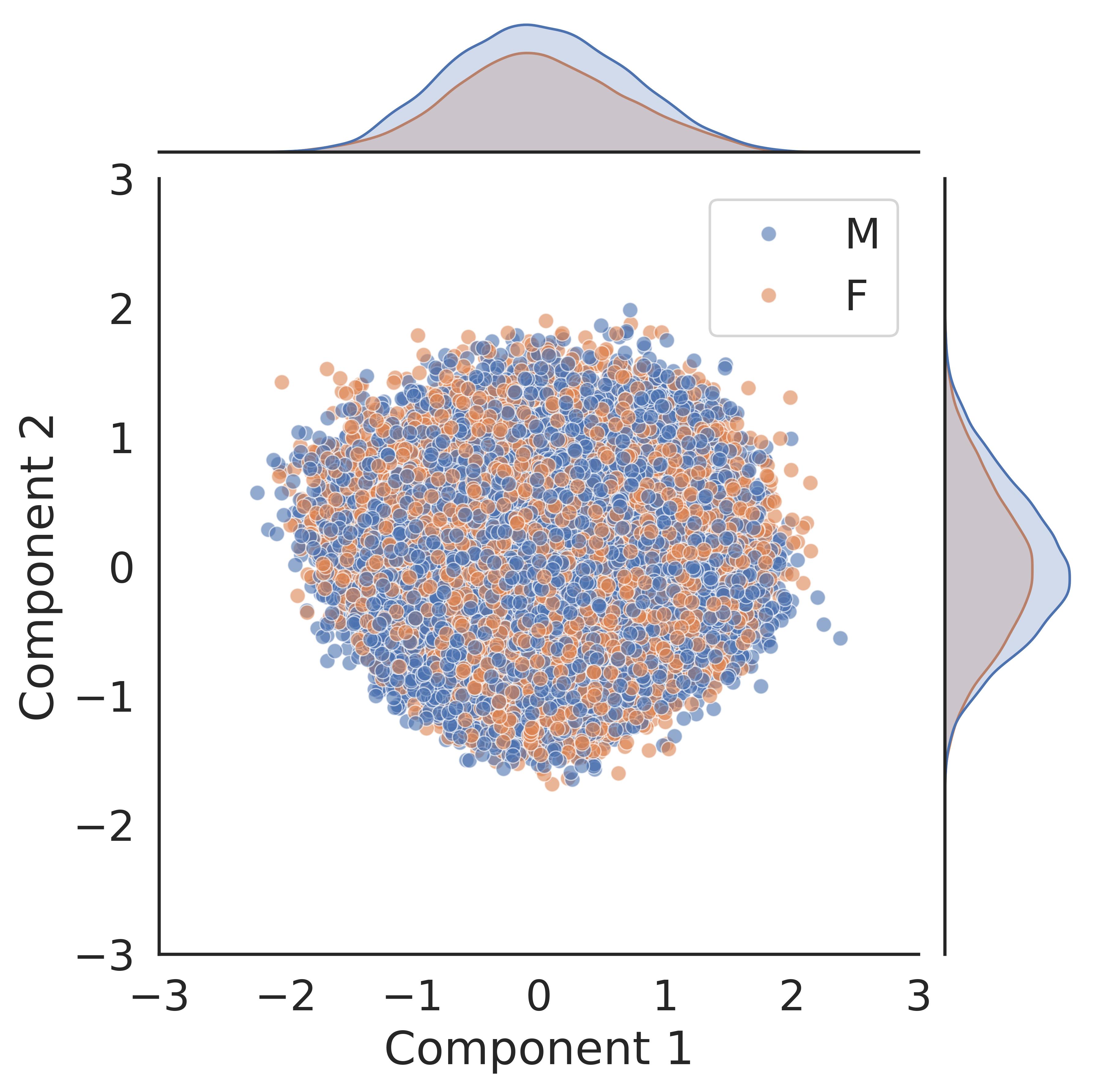} &
        \includegraphics[width=.20\textwidth]{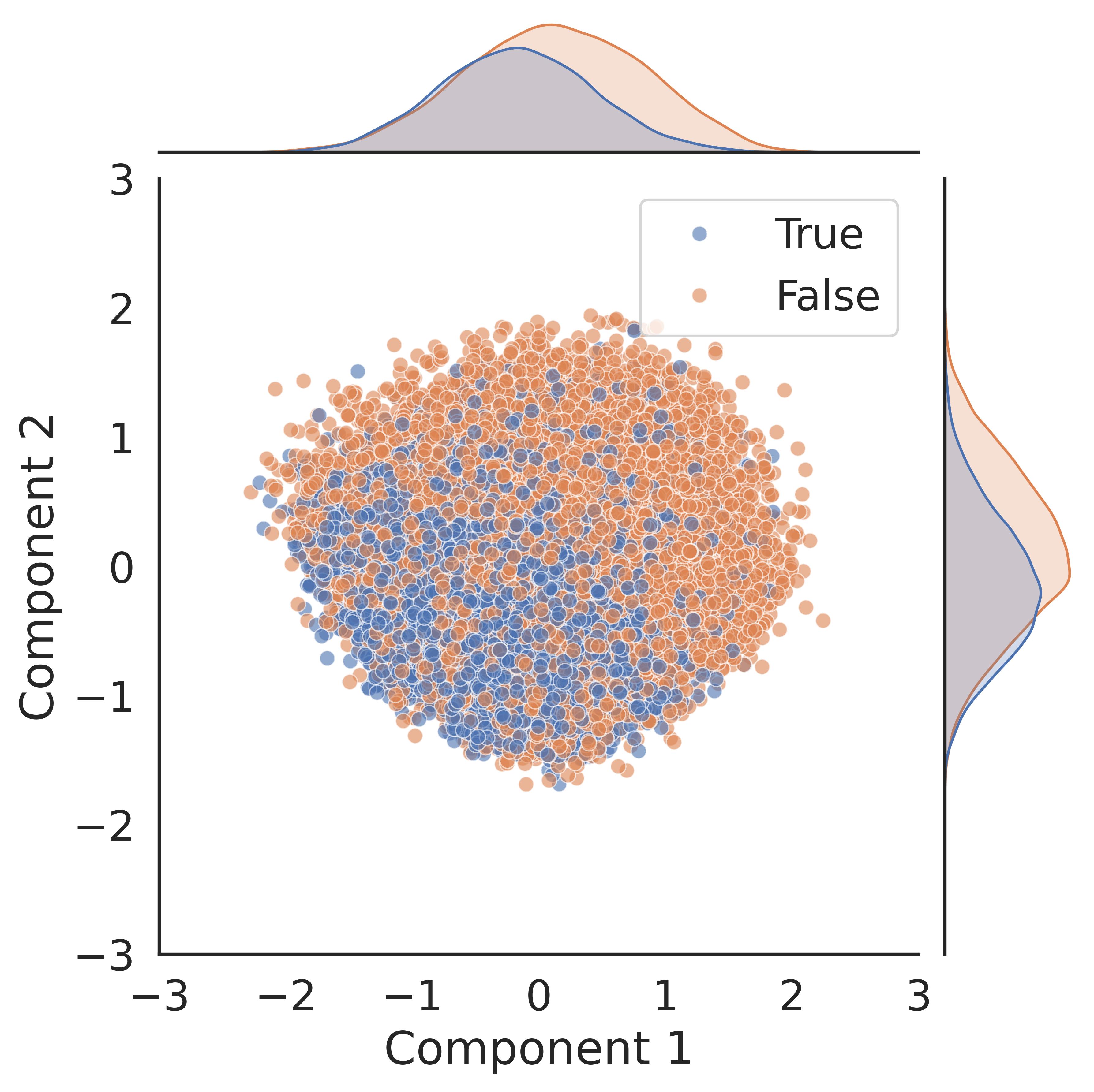} &
        \includegraphics[width=.20\textwidth]{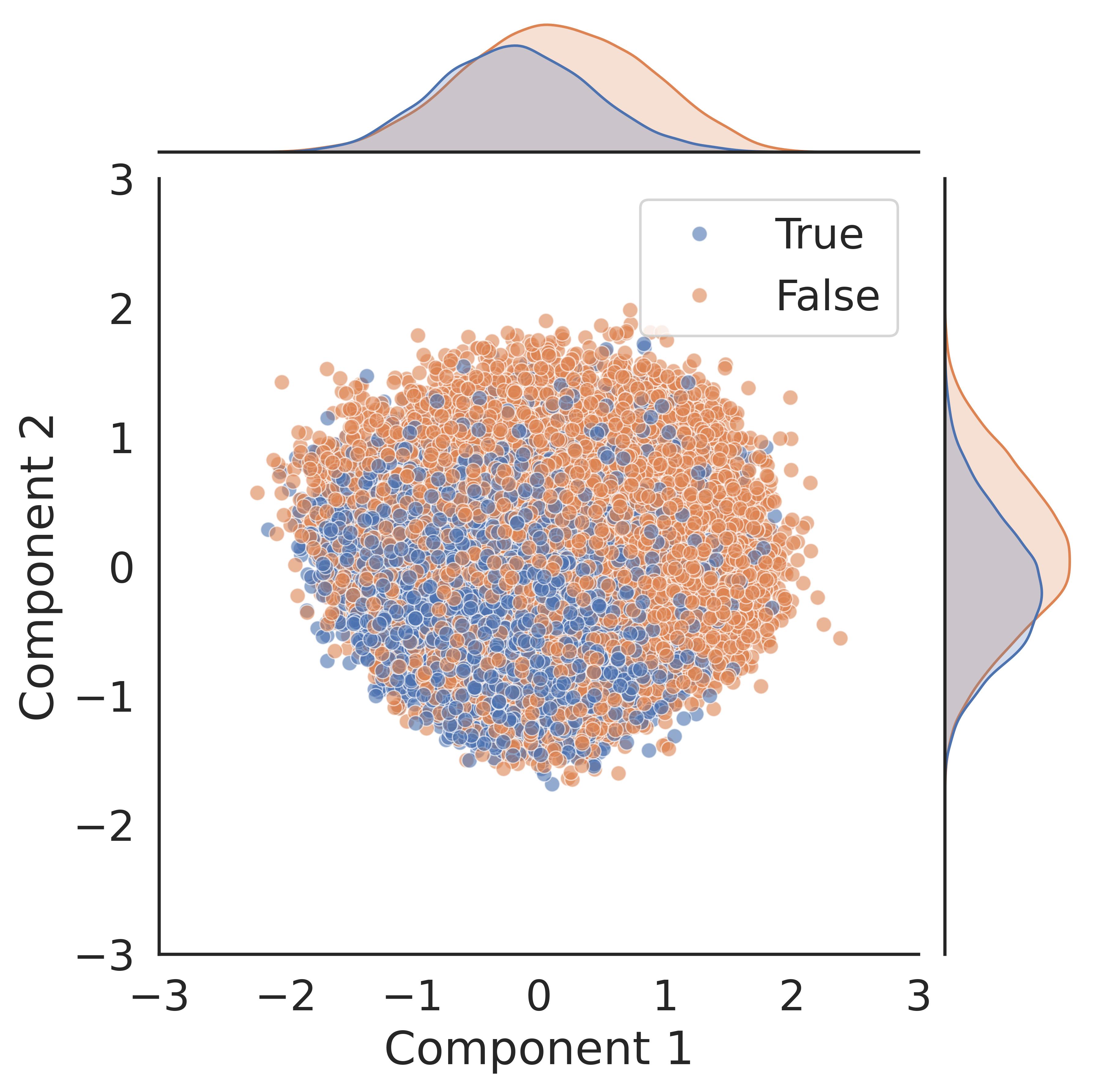} \\
        \midrule
        \rotatebox[origin=l]{90}{\textbf{CheXpert CLF}} &
        \includegraphics[width=.20\textwidth]{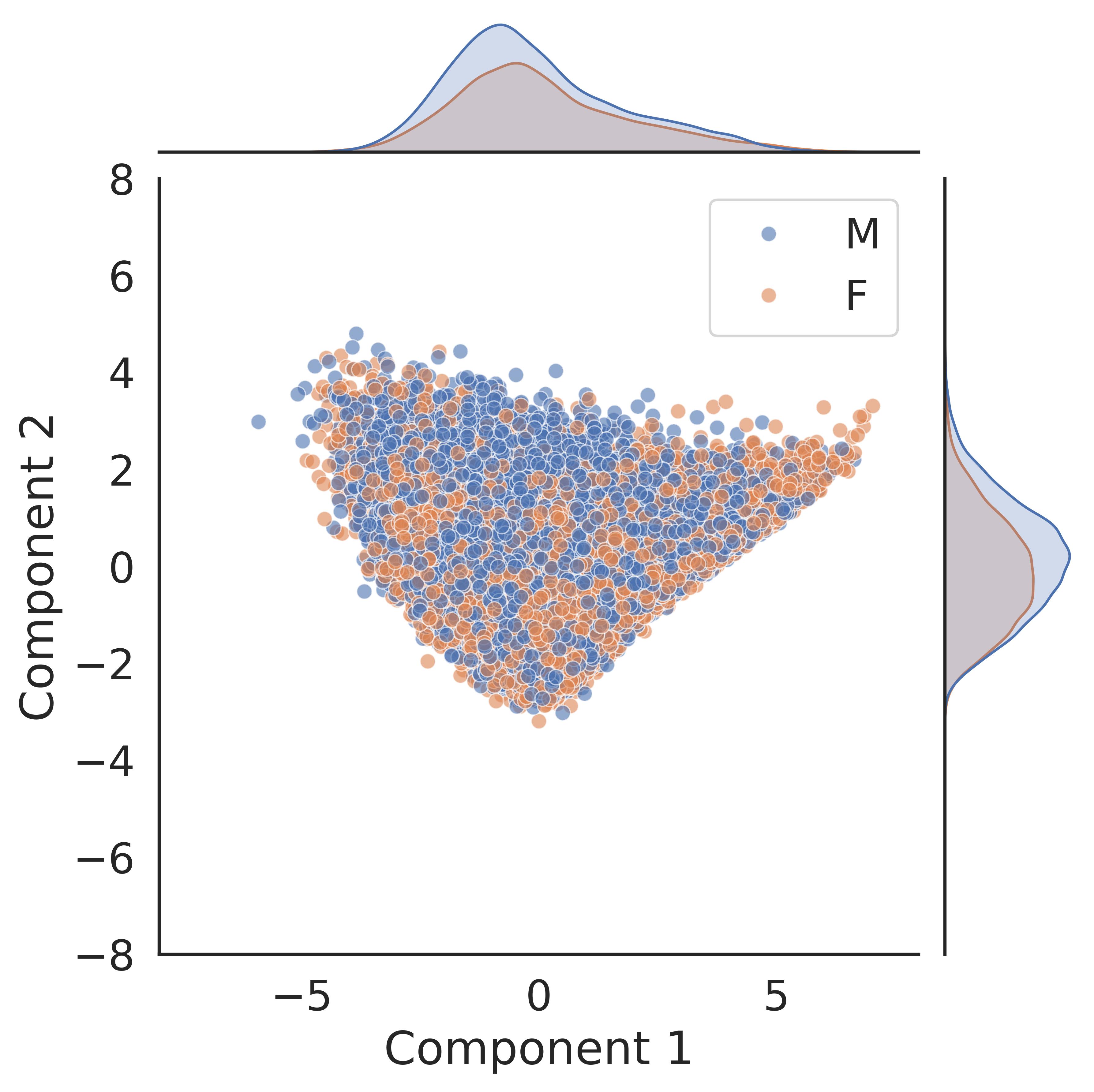} &
        \includegraphics[width=.20\textwidth]{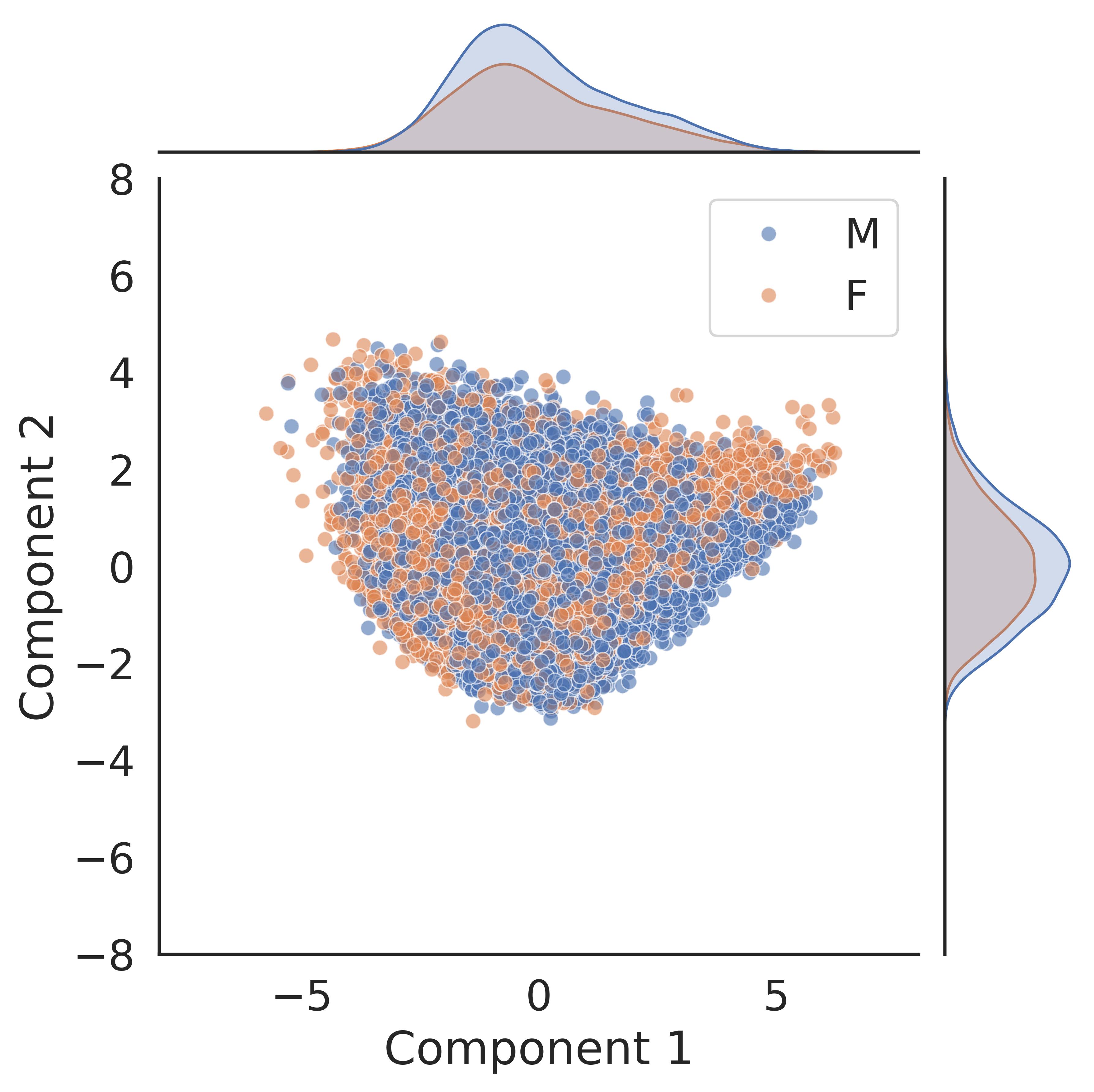} &
        \includegraphics[width=.20\textwidth]{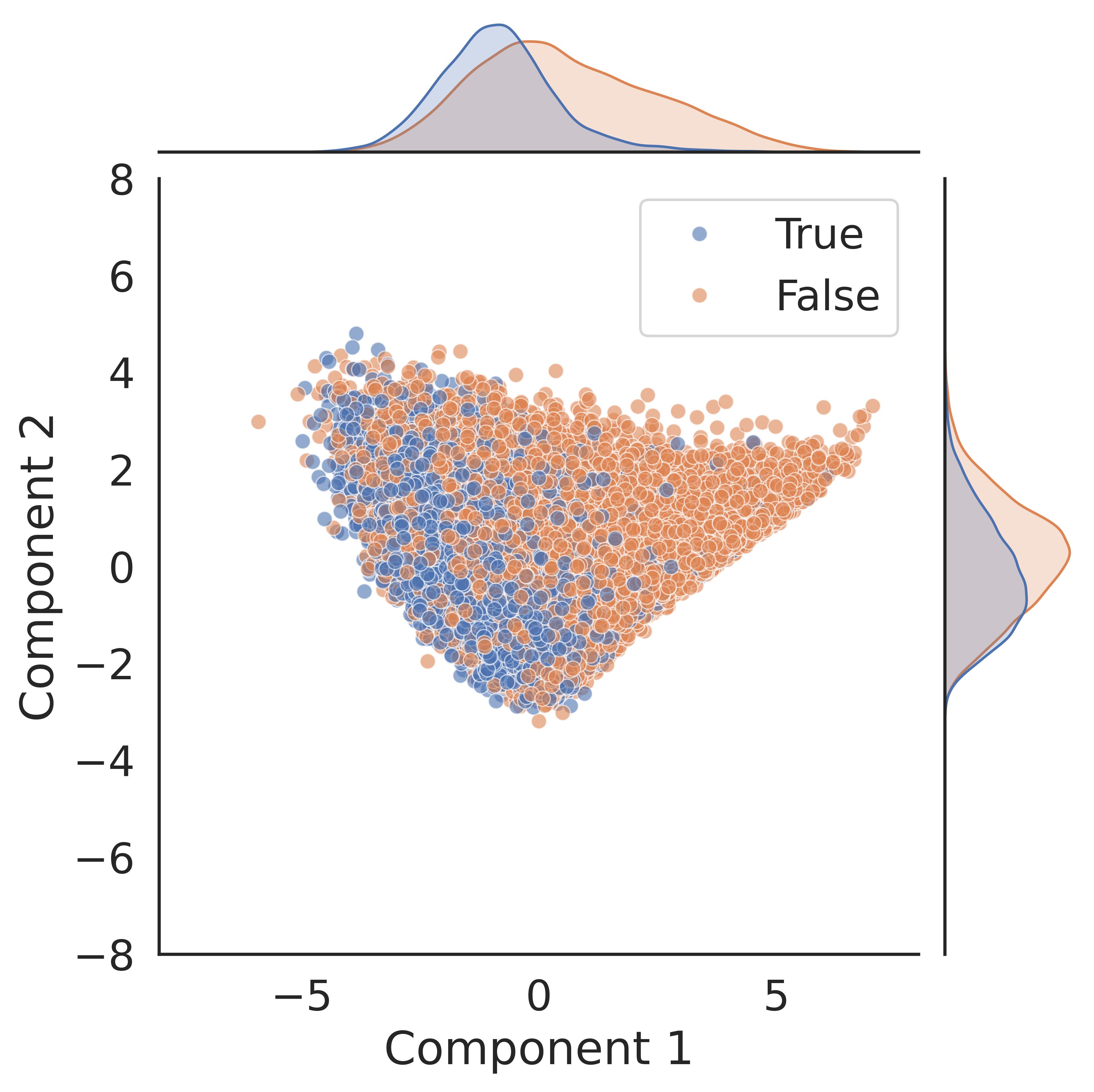} &
        \includegraphics[width=.20\textwidth]{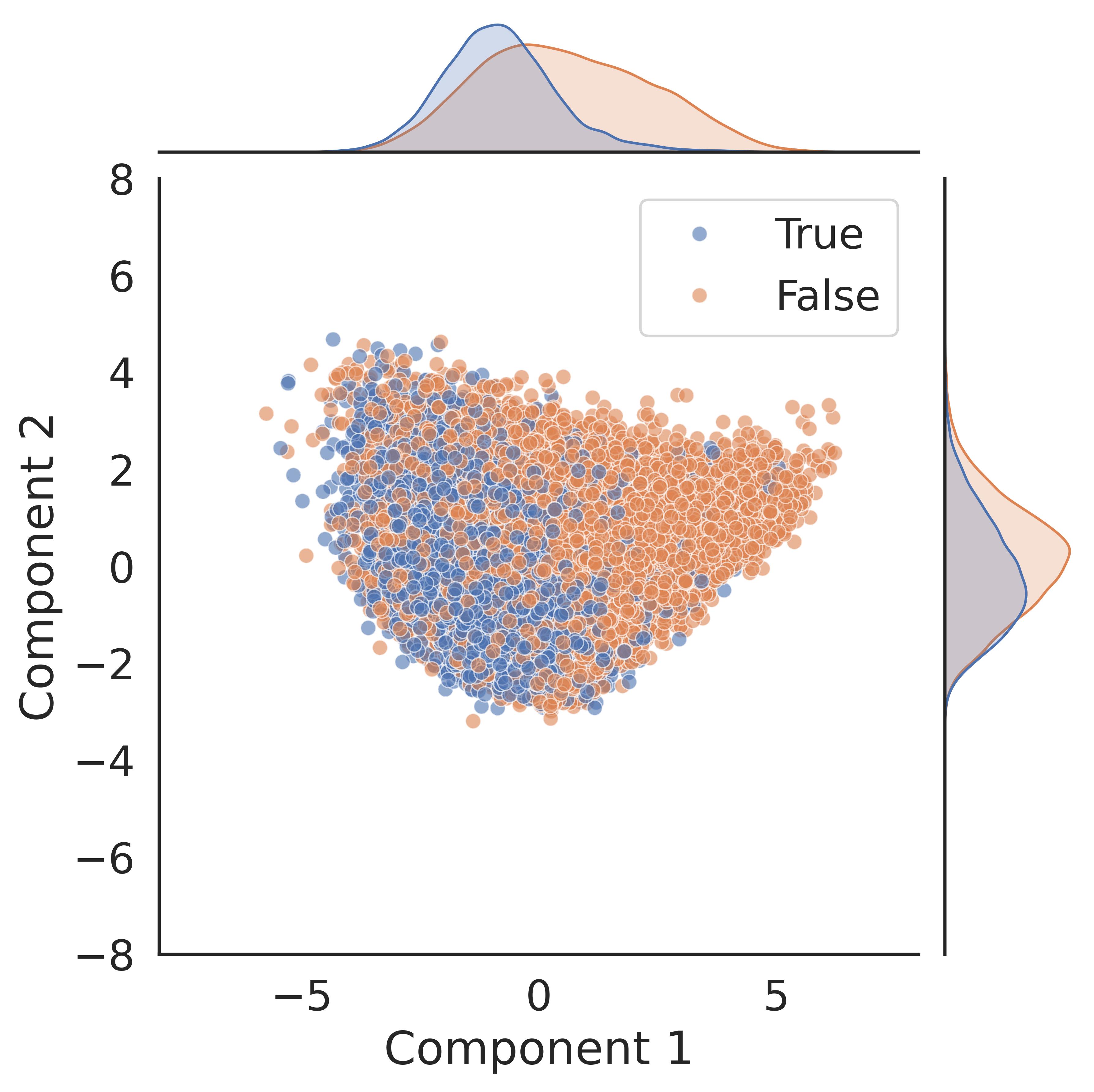} \\
    \end{tabular}
    }
    \caption{Marginalized distributions per sex (\textbf{left} two columns) and label \textit{Pleural Effusion} (\textbf{right} two columns) over the PCA reduction of the original versus the orthogonalized corrected embedding.}
    \label{fig:pca-combined}
\end{figure}

\paragraph{With Orthogonalization}
After applying orthogonalization, the resulting regression coefficients are estimated to be zero and exhibit p-values of one, implying that the protected feature information is successfully removed in the corrected embedding.

Figure~\ref{fig:pca-combined} further shows the effect of the orthogonalization on the first two PCA components of the embedding.
The figure visualizes the marginal distributions of these two components for the two available sex labels (M/F) across all evaluated modalities before and after orthogonalization.
The orthogonalization procedure does not change the directions of the highest variance in the datasets and preserves the general structure of the embedding space.
However, when examining the differences in the two sex labels, we notice an alignment in the marginal distributions.
This is particularly striking for the marginals of the second principal component in the MIMIC CFM and CheSS embeddings. After applying orthogonalization, the shift between those distributions is removed, resulting in a loss of discriminative power and elimination of bias regarding this feature.
In contrast, there are no or only minor visual changes in the marginals of the first two PCA components for \textit{Pleural Effusion} (cf.~Figure~\ref{fig:pca-combined}).

\subsection{Predicting Protected Information}

\begin{figure}[t]
  \centering
  \begin{subfigure}{0.33\textwidth}
    \centering
    \includegraphics[width=\linewidth]{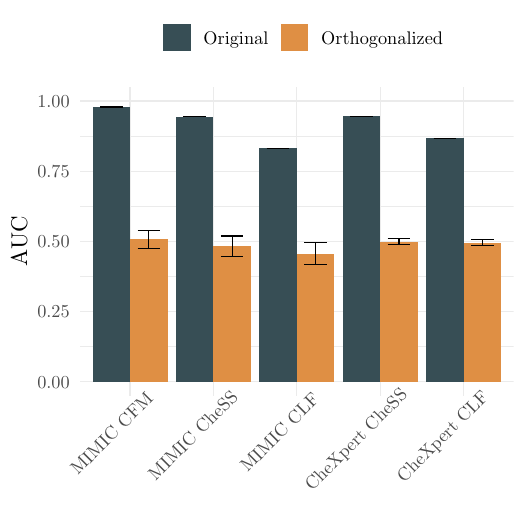}
    \caption{Classification of sex.}
  \end{subfigure}%
  \begin{subfigure}{0.33\textwidth}
    \centering
    \includegraphics[width=\linewidth]{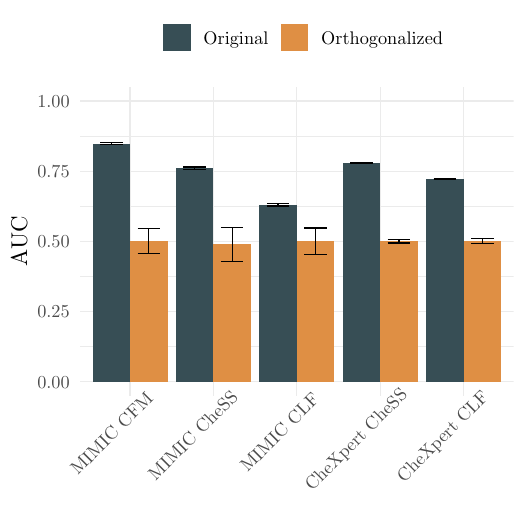}
    \caption{Classification of \textit{White}.}
  \end{subfigure}%
  \begin{subfigure}{0.33\textwidth}
    \centering
    \includegraphics[width=\linewidth]{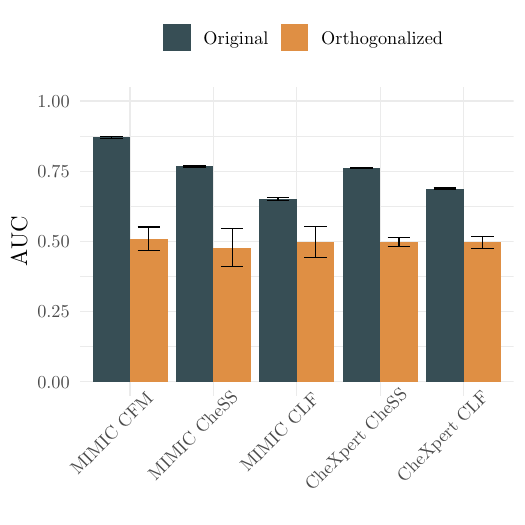}
    \caption{Classification of \textit{Black}.}
  \end{subfigure}
  \vspace{1em}
  \\ 
  \begin{subfigure}{0.33\textwidth}
    \centering
    \includegraphics[width=\linewidth]{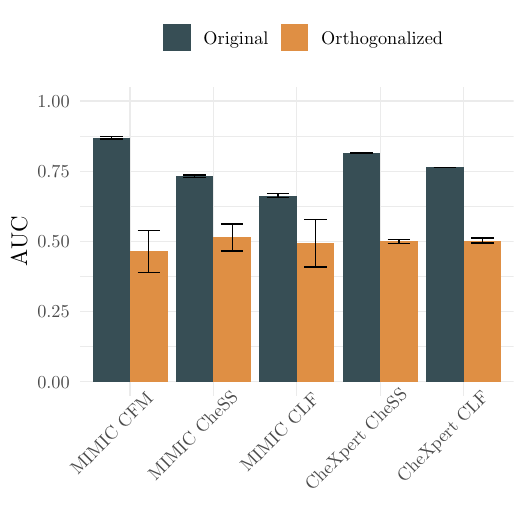}
    \caption{Classification of \textit{Asian}.}
  \end{subfigure}%
  \begin{subfigure}{0.33\textwidth}
    \centering
    \includegraphics[width=\linewidth]{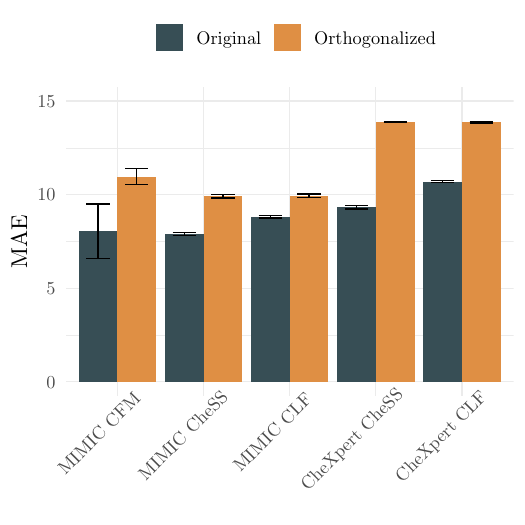}
    \caption{Regression of age.}
  \end{subfigure}

    \caption{Regression/Classification performance for deriving protected features from an embedding vector with mean and standard deviation over 10 randomly initialized runs. The displayed metrics include mean absolute error (MAE) in years for age regression as well as AUC for classification of sex and race.}
    \label{fig:feat-reg}
\end{figure}

Apart from their influence on pathology classification, we investigate how well the protected information can be predicted directly from the embeddings before and after orthogonalization.
As depicted in Figure~\ref{fig:feat-reg} with extended metrics in Table~\ref{tab:class-protected}, identification of protected features in deep embeddings can be achieved with high accuracy, e.g., a patient's sex with an AUC of 0.979 in MIMIC CFM.
With orthogonalization, a classification is not feasible anymore and results in a mean AUC of $0.507$, equivalent to random guessing.
This observation holds equally for the prediction of race, as well as for the prediction of a patient's age.
In the case of CheXpert`s CheSS embeddings, linear regression yields an $R^2$ of $0.529$, whereas, after correction, the $R^2$ is $-0.009$.
Generally, the two CLF models show the lowest AUCs when predicting sex or race.

\subsection{Downstream Prediction Performance} \label{sec:res-down}

Lastly, another important question is how orthogonalization influences embeddings with respect to their predictive power.
Table~\ref{tab:auc-scores} and Supplement~\ref{tab:pred-scores} contain an overview of the estimated downstream prediction performances when training a logistic regression on both, the original and corrected embeddings, separately.
The MIMIC CFM embedding has a notably higher macro-averaged AUC than CheSS and CLF ($0.789$ versus $0.695$ and $0.679$.)
On CheXpert, the CLF embedding with an AUC of $0.720$ yields a slightly better result than CheSS with $0.710$.
For CheSS and CLF, the difference in performance is $-0.14\%$ , while for the CFM embeddings, the decrease is $-1.39\%$.
However, in the case of CheXpert, the removal of the protected feature even results in a performance boost, e.g., $+1.83\%$ for CheSS.
\begin{figure}[b!]
  \begin{subfigure}{0.40\textwidth}
    \centering
    \includegraphics[width=\linewidth]{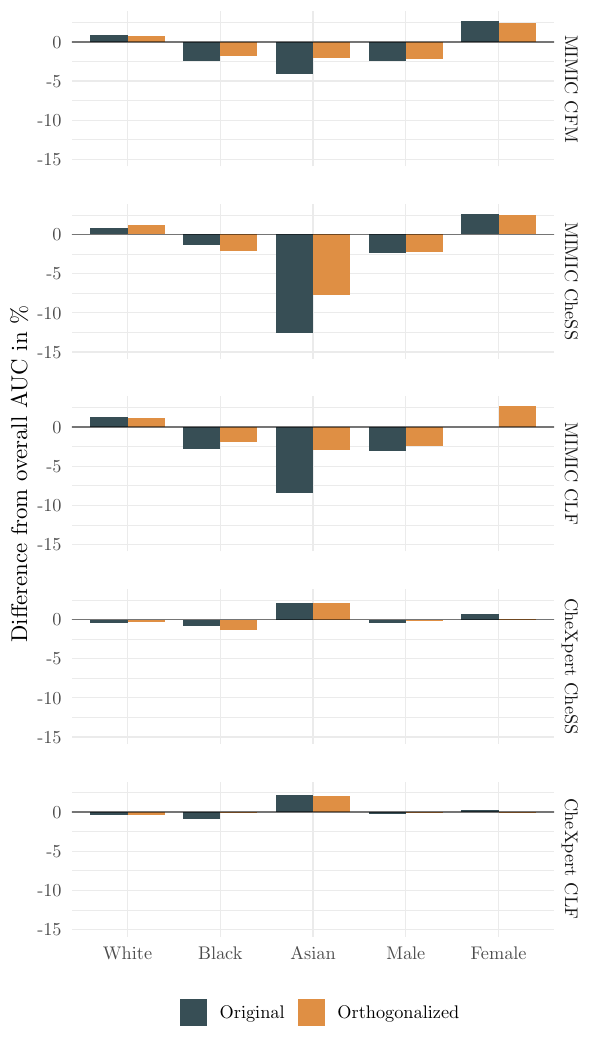}
    \caption{Change in AUC for \textit{Pleural Effusion}.}
  \end{subfigure}\hfill
    %
  \begin{subfigure}{0.40\textwidth}
    \centering
    \includegraphics[width=\linewidth]{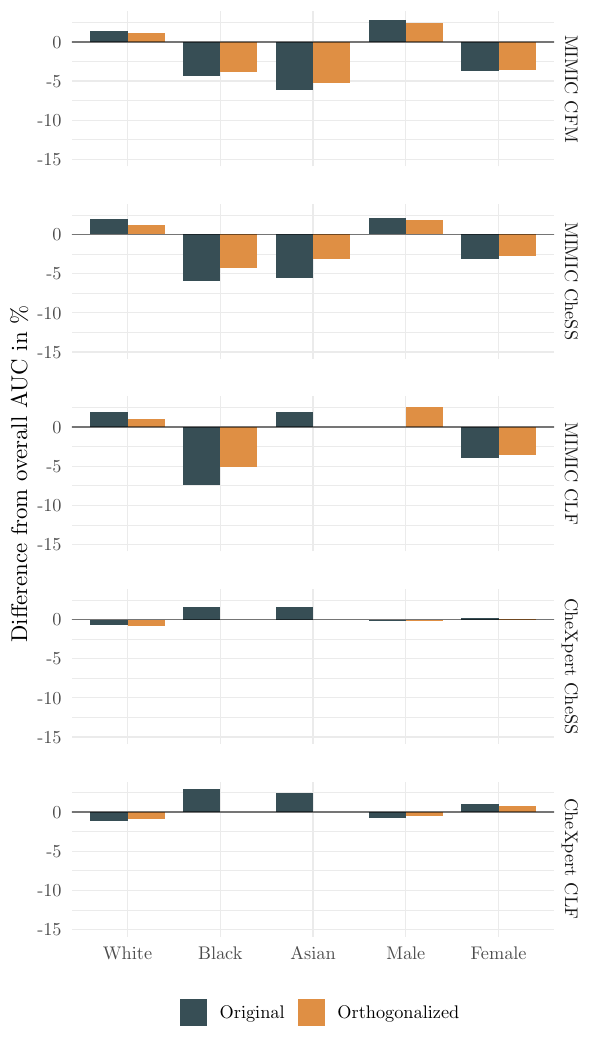}
    \caption{Change in AUC for \textit{Cardiomegaly}.}
  \end{subfigure}
    \caption{Difference from overall AUC per subgroup based on downstream classifiers for original and orthogonalized embeddings. The closer a bar is to zero, the less disparity from the mean AUC exists.}
    \label{fig:auc-difference-subgroup}
\end{figure}

Figure~\ref{fig:auc-difference-subgroup} depicts the deviation of the subgroup AUC from the overall AUC for \textit{Pleural Effusion} and \textit{Cardiomegaly}.
Using the orthogonalization, in fact, mitigates the disparity in most cases, e.g., the \textit{Pleural Effusion} class in MIMIC Chess from $-12.54\%$ to $-7.76\%$.

\begin{landscape}
    \begin{table}[t]
    \centering
    \resizebox{1.6\textwidth}{!}{%
    \begin{tabular}{llccc | ccc | ccc | ccc | ccc}
    \toprule
    & & & \multicolumn{2}{c}{\textbf{Age}} & \multicolumn{3}{c}{\textbf{Sex}} & \multicolumn{3}{c}{\textbf{Race [White]}} & \multicolumn{3}{c}{\textbf{Race [Black]}} & \multicolumn{3}{c}{\textbf{Race [Asian]}} \\
    \cline{4-5} \cline{6-8} \cline{9-11} \cline{12-14} \cline{15-17}
    & \textbf{Emb.} & \textbf{Orthogonalized?} & \textbf{MAE} & $\bm{R}^{\bm 2}$ & \textbf{AUC} & \textbf{Sens.} & \textbf{Spec.} & \textbf{AUC} & \textbf{Sens.} & \textbf{Spec.} & \textbf{AUC} & \textbf{Sens.} & \textbf{Spec.} & \textbf{AUC} & \textbf{Sens.} & \textbf{Spec.}\\
    \midrule
    \multirow{6}{*}{\rotatebox[origin=c]{90}{\textbf{MIMIC}}}
    & \multirow{2}{*}{CFM} & \xmark     & 8.040 $\pm$1.456 & 0.314 $\pm$0.241  & 0.979 $\pm$0.000 & 0.933 $\pm$0.025 & 0.902 $\pm$0.039 & 0.848 $\pm$0.004 & 0.940 $\pm$0.016 & 0.497 $\pm$0.049 & 0.871 $\pm$0.004 & 0.521 $\pm$0.055 & 0.944 $\pm$0.015 & 0.870 $\pm$0.005 & 0.179 $\pm$0.028 & 0.991 $\pm$0.005 \\
    &                      & \cmark     & 10.969 $\pm$0.433 & -0.294 $\pm$0.087& 0.507 $\pm$0.031 & 0.775 $\pm$0.077 & 0.224 $\pm$0.100 & 0.501 $\pm$0.045 & 1.000 $\pm$0.000 & 0.000 $\pm$0.000 & 0.509 $\pm$0.042 & 0.000 $\pm$0.000 & 1.000 $\pm$0.000 & 0.464 $\pm$0.076 & 0.000 $\pm$0.000 & 1.000 $\pm$0.000 \\
    \cmidrule{2-17}
    & \multirow{2}{*}{CheSS} & \xmark   & 7.893 $\pm$ 0.066 & 0.331 $\pm$ 0.010 & 0.945 $\pm$ 0.001 & 0.907 $\pm$ 0.011 & 0.820 $\pm$ 0.016 & 0.761 $\pm$ 0.004 & 0.975 $\pm$ 0.008 & 0.158 $\pm$ 0.039 & 0.768 $\pm$ 0.003 & 0.143 $\pm$ 0.039 & 0.968 $\pm$ 0.010 & 0.733 $\pm$ 0.004 & 0.013 $\pm$ 0.008 & 0.997 $\pm$ 0.002 \\
    &                      & \cmark     & 9.908 $\pm$ 0.082 & -0.083 $\pm$ 0.016& 0.482 $\pm$ 0.037 & 0.982 $\pm$ 0.024 & 0.013 $\pm$ 0.025 & 0.489 $\pm$ 0.060 & 1.000 $\pm$ 0.000 & 0.000 $\pm$ 0.000 & 0.478 $\pm$ 0.068 & 0.000 $\pm$ 0.000 & 1.000 $\pm$ 0.000 & 0.514 $\pm$ 0.048 & 0.000 $\pm$ 0.000 & 1.000 $\pm$ 0.000 \\
    \cmidrule{2-17}
    & \multirow{2}{*}{CLF} & \xmark     & 8.816 $\pm$ 0.063 & 0.161 $\pm$ 0.011 & 0.832 $\pm$ 0.000 & 0.804 $\pm$ 0.015 & 0.702 $\pm$ 0.016 & 0.631 $\pm$ 0.005 & 0.988 $\pm$ 0.004 & 0.048 $\pm$ 0.012 & 0.652 $\pm$ 0.005 & 0.050 $\pm$ 0.012 & 0.987 $\pm$ 0.005 & 0.663 $\pm$ 0.007 & 0.000 $\pm$ 0.000 & 1.000 $\pm$ 0.000 \\
    &                      & \cmark     & 9.941 $\pm$ 0.095 & -0.093 $\pm$ 0.025& 0.456 $\pm$ 0.039 & 0.994 $\pm$ 0.007 & 0.007 $\pm$ 0.006 & 0.500 $\pm$ 0.048 & 1.000 $\pm$ 0.000 & 0.000 $\pm$ 0.000 & 0.498 $\pm$ 0.056 & 0.000 $\pm$ 0.000 & 1.000 $\pm$ 0.000 & 0.493 $\pm$ 0.084 & 0.000 $\pm$ 0.000 & 1.000 $\pm$ 0.000 \\
    \midrule
    \multirow{4}{*}{\rotatebox[origin=c]{90}{\textbf{CheXpert}}}
    & \multirow{2}{*}{CheSS} & \xmark   & 9.333 $\pm$ 0.103 & 0.529 $\pm$ 0.009 & 0.946 $\pm$ 0.000 & 0.912 $\pm$ 0.008 & 0.821 $\pm$ 0.018 & 0.780 $\pm$ 0.001 & 0.984 $\pm$ 0.007 & 0.136 $\pm$ 0.038 & 0.762 $\pm$ 0.001 & 0.018 $\pm$ 0.006 & 0.999 $\pm$ 0.001 & 0.816 $\pm$ 0.001 & 0.173 $\pm$ 0.047 & 0.983 $\pm$ 0.007 \\
    &                      & \cmark     &13.875 $\pm$ 0.035 & -0.009 $\pm$ 0.005& 0.499 $\pm$ 0.010 & 1.000 $\pm$ 0.000 & 0.000 $\pm$ 0.000 & 0.501 $\pm$ 0.007 & 1.000 $\pm$ 0.000 & 0.000 $\pm$ 0.000 & 0.498 $\pm$ 0.015 & 0.000 $\pm$ 0.000 & 1.000 $\pm$ 0.000 & 0.500 $\pm$ 0.007 & 0.000 $\pm$ 0.000 & 1.000 $\pm$ 0.000 \\
    \cmidrule{2-17}
    & \multirow{2}{*}{CLF} & \xmark     & 10.693 $\pm$ 0.047 & 0.385 $\pm$ 0.005&0.868 $\pm$ 0.000 & 0.854 $\pm$ 0.008 & 0.696 $\pm$ 0.014 & 0.721 $\pm$ 0.001 & 0.993 $\pm$ 0.003 & 0.040 $\pm$ 0.012 & 0.688 $\pm$ 0.002 & 0.003 $\pm$ 0.002 & 1.000 $\pm$ 0.000 & 0.764 $\pm$ 0.000 & 0.049 $\pm$ 0.016 & 0.993 $\pm$ 0.003 \\
    &                      & \cmark     &13.866 $\pm$ 0.035 & -0.009 $\pm$ 0.005&0.496 $\pm$ 0.010 & 1.000 $\pm$ 0.000 & 0.000 $\pm$ 0.000 & 0.501 $\pm$ 0.008 & 1.000 $\pm$ 0.000 & 0.000 $\pm$ 0.000 & 0.497 $\pm$ 0.021 & 0.000 $\pm$ 0.000 & 1.000 $\pm$ 0.000 & 0.503 $\pm$ 0.009 & 0.000 $\pm$ 0.000 & 1.000 $\pm$ 0.000\\
    \bottomrule
    \end{tabular}
    }
    \caption{Regression/Classification performance for deriving protected features from an embedding vector with mean and standard deviation over 10 randomly initialized runs. The displayed metrics include mean absolute error (MAE), $R^2$ for age regression as well as AUC, sensitivity (sens.), and specificity (spec.) for classification.}
    \label{tab:class-protected}
\end{table}
\begin{table}[t]
    \centering
    \resizebox{1.6\textwidth}{!}{%
    \begin{tabular}{lccc|ccc|ccc|ccc|ccc}
    \toprule
    \textbf{Dataset:} & \multicolumn{9}{c}{\textbf{MIMIC}} & \multicolumn{6}{c}{\textbf{CheXpert}} \\
    \cline{2-11} \cline{12-16}
    \textbf{Embedding:} & \multicolumn{3}{c}{\textbf{CFM}} & \multicolumn{3}{c}{\textbf{CheSS}} & \multicolumn{3}{c}{\textbf{CLF}} & \multicolumn{3}{c}{\textbf{CheSS}} & \multicolumn{3}{c}{\textbf{CLF}}\\
    \midrule
    \textbf{Orthogonalized?} & \xmark & \cmark & $\Delta$ & \xmark & \cmark & $\Delta$ & \xmark & \cmark & $\Delta$ & \xmark & \cmark & $\Delta$ & \xmark & \cmark & $\Delta$ \\
    \textbf{Enl. Cardiomed.}  & 0.728 $\pm$ 0.009 & 0.721 $\pm$ 0.018 & -0.97 \% & 0.636 $\pm$ 0.003 & 0.643 $\pm$ 0.007 & +1.09 \% & 0.601 $\pm$ 0.002 & 0.593 $\pm$ 0.004 & -1.35 \% & 0.621 $\pm$ 0.001 & 0.636 $\pm$ 0.003 & +2.36 \% & 0.634 $\pm$ 0.000 & 0.639 $\pm$ 0.001 & +0.78 \% \\
    \textbf{Cardiomegaly}     & 0.780 $\pm$ 0.002 & 0.775 $\pm$ 0.003 & -0.65 \% & 0.750 $\pm$ 0.001 & 0.751 $\pm$ 0.001 & +0.13 \% & 0.737 $\pm$ 0.000 & 0.736 $\pm$ 0.001 & -0.14 \% & 0.789 $\pm$ 0.000 & 0.791 $\pm$ 0.001 & +0.25 \% & 0.799 $\pm$ 0.000 & 0.793 $\pm$ 0.000 & -0.76 \% \\
    \textbf{Lung Opacity}     & 0.696 $\pm$ 0.003 & 0.684 $\pm$ 0.005 & -1.75 \% & 0.626 $\pm$ 0.002 & 0.627 $\pm$ 0.004 & +0.16 \% & 0.623 $\pm$ 0.001 & 0.612 $\pm$ 0.002 & -1.80 \% & 0.685 $\pm$ 0.000 & 0.684 $\pm$ 0.000 & -0.15 \% & 0.695 $\pm$ 0.000 & 0.690 $\pm$ 0.000 & -0.72 \% \\
    \textbf{Lung Lesion}      & 0.731 $\pm$ 0.006 & 0.718 $\pm$ 0.007 & -1.81 \% & 0.623 $\pm$ 0.003 & 0.630 $\pm$ 0.004 & +1.11 \% & 0.576 $\pm$ 0.009 & 0.591 $\pm$ 0.014 & +2.54 \% & 0.672 $\pm$ 0.002 & 0.700 $\pm$ 0.002 & +4.00 \% & 0.701 $\pm$ 0.001 & 0.707 $\pm$ 0.002 & +0.85 \% \\
    \textbf{Edema}            & 0.843 $\pm$ 0.001 & 0.837 $\pm$ 0.002 & -0.72 \% & 0.804 $\pm$ 0.000 & 0.798 $\pm$ 0.000 & -0.75 \% & 0.791 $\pm$ 0.000 & 0.783 $\pm$ 0.001 & -1.02 \% & 0.791 $\pm$ 0.000 & 0.789 $\pm$ 0.000 & -0.25 \% & 0.788 $\pm$ 0.000 & 0.783 $\pm$ 0.000 & -0.64 \% \\
    \textbf{Consolidation}    & 0.748 $\pm$ 0.008 & 0.742 $\pm$ 0.009 & -0.81 \% & 0.648 $\pm$ 0.001 & 0.650 $\pm$ 0.005 & +0.31 \% & 0.638 $\pm$ 0.003 & 0.640 $\pm$ 0.002 & +0.31 \% & 0.669 $\pm$ 0.001 & 0.683 $\pm$ 0.001 & +2.05 \% & 0.689 $\pm$ 0.001 & 0.692 $\pm$ 0.002 & +0.43 \% \\
    \textbf{Pneumonia}        & 0.703 $\pm$ 0.005 & 0.704 $\pm$ 0.004 & +0.14 \% & 0.586 $\pm$ 0.005 & 0.597 $\pm$ 0.009 & +1.84 \% & 0.589 $\pm$ 0.003 & 0.607 $\pm$ 0.002 & +2.97 \% & 0.610 $\pm$ 0.002 & 0.652 $\pm$ 0.003 & +6.44 \% & 0.652 $\pm$ 0.001 & 0.659 $\pm$ 0.004 & +1.06 \% \\
    \textbf{Atelectasis}      & 0.746 $\pm$ 0.002 & 0.734 $\pm$ 0.004 & -1.63 \% & 0.702 $\pm$ 0.000 & 0.696 $\pm$ 0.001 & -0.86 \% & 0.685 $\pm$ 0.001 & 0.671 $\pm$ 0.001 & -2.09 \% & 0.631 $\pm$ 0.000 & 0.636 $\pm$ 0.001 & +0.79 \% & 0.632 $\pm$ 0.000 & 0.633 $\pm$ 0.001 & +0.16 \% \\
    \textbf{Pneumothorax}     & 0.843 $\pm$ 0.005 & 0.830 $\pm$ 0.007 & -1.57 \% & 0.649 $\pm$ 0.002 & 0.645 $\pm$ 0.005 & -0.62 \% & 0.634 $\pm$ 0.003 & 0.638 $\pm$ 0.004 & +0.63 \% & 0.732 $\pm$ 0.001 & 0.749 $\pm$ 0.001 & +2.27 \% & 0.730 $\pm$ 0.001 & 0.740 $\pm$ 0.001 & +1.35 \% \\
    \textbf{Pleural Effusion} & 0.870 $\pm$ 0.001 & 0.859 $\pm$ 0.002 & -1.28 \% & 0.802 $\pm$ 0.000 & 0.792 $\pm$ 0.001 & -1.26 \% & 0.797 $\pm$ 0.000 & 0.781 $\pm$ 0.000 & -2.05 \% & 0.792 $\pm$ 0.000 & 0.798 $\pm$ 0.000 & +0.75 \% & 0.804 $\pm$ 0.000 & 0.801 $\pm$ 0.000 & -0.37 \% \\
    \textbf{Pleural Other}    & 0.894 $\pm$ 0.009 & 0.874 $\pm$ 0.021 & -2.29 \% & 0.711 $\pm$ 0.005 & 0.746 $\pm$ 0.011 & +4.69 \% & 0.684 $\pm$ 0.006 & 0.693 $\pm$ 0.009 & +1.30 \% & 0.723 $\pm$ 0.001 & 0.756 $\pm$ 0.004 & +4.37 \% & 0.718 $\pm$ 0.001 & 0.718 $\pm$ 0.004 & +0.00 \% \\
    \textbf{Fracture}         & 0.752 $\pm$ 0.007 & 0.739 $\pm$ 0.013 & -1.76 \% & 0.643 $\pm$ 0.005 & 0.648 $\pm$ 0.009 & +0.77 \% & 0.642 $\pm$ 0.004 & 0.652 $\pm$ 0.012 & +1.53 \% & 0.668 $\pm$ 0.001 & 0.682 $\pm$ 0.003 & +2.05 \% & 0.667 $\pm$ 0.001 & 0.673 $\pm$ 0.003 & +0.89 \% \\
    \textbf{Support Devices}  & 0.909 $\pm$ 0.001 & 0.905 $\pm$ 0.001 & -0.44 \% & 0.801 $\pm$ 0.000 & 0.801 $\pm$ 0.001 & +0.00 \% & 0.767 $\pm$ 0.001 & 0.763 $\pm$ 0.001 & -0.52 \% & 0.731 $\pm$ 0.000 & 0.748 $\pm$ 0.000 & +2.27 \% & 0.711 $\pm$ 0.000 & 0.721 $\pm$ 0.000 & +1.39 \% \\
    \textbf{No Finding}       & 0.801 $\pm$ 0.003 & 0.770 $\pm$ 0.005 & -4.03 \% & 0.746 $\pm$ 0.001 & 0.725 $\pm$ 0.002 & -2.90 \% & 0.747 $\pm$ 0.000 & 0.728 $\pm$ 0.001 & -2.61 \% & 0.833 $\pm$ 0.000 & 0.824 $\pm$ 0.001 & -1.09 \% & 0.854 $\pm$ 0.000 & 0.844 $\pm$ 0.000 & -1.18 \% \\
    \bottomrule
    \textbf{Total}       & 0.789 $\pm$ 0.005 & 0.778 $\pm$ 0.009 & -1.39 \% & 0.695 $\pm$ 0.003 & 0.696 $\pm$ 0.005 & -0.14 \% & 0.679 $\pm$ 0.003 & 0.678 $\pm$ 0.006 & -0.14 \% & 0.710 $\pm$ 0.001 & 0.723 $\pm$ 0.002 & +1.83 \% & 0.720 $\pm$ 0.001 & 0.721 $\pm$ 0.002 & +0.13 \% \\
    \bottomrule
    \end{tabular}
    }
    \caption{Prediction performance original versus orthogonalized data on the MIMIC and CheXpert datasets. The table shows the mean and standard deviation of the AUC over 10 randomly initialized runs. Additionally, $\Delta$ depicts the percentual change from the original to the corrected embedding AUC.}
    \label{tab:auc-scores}
\end{table}
\end{landscape}
%

\section{Discussion}

Our analysis investigates the implicit encoding of protected features age, sex, and race in three independent neural representations of MIMIC and CheXpert, and shows a significant influence of these protected features on model predictions.
Using an orthogonalization allows us to remove this effect.
Our findings indicate that recovering protected features from orthogonalized embeddings is not possible anymore and their influence on predictions is eliminated.
Orthogonalization further allows the reduction of subgroup disparities.

The phenomenon of bias in CXR classifiers is a highly active research area.
\cite{gichoya2023ai} identify various sources of bias during all stages of model development and
\cite{banerjee2023shortcuts} account shortcut learning as a possible origin of such biases.
Further, \cite{seyyed2021underdiagnosis} detect an underdiagnosis bias in underrepresented patient populations.
\cite{glocker2023algorithmic} warn about encoded protected features in latent representations of classifiers.
More in-depth analyses about subgroup disparities are conducted in \cite{seyyed2020chexclusion, ahluwalia2023subgroup}.
We contribute to this discussion by not only confirming the existence of biases but also by proposing additional measures to effectively remove them.

Previous work \cite{glocker2022risk} already addressed potential biases in the CheXpert embeddings of CFM by testing for differences in the mean of specific subgroups in the first four PCA components.
As this comes with the downside of evaluating only one protected feature isolated at a time and does not directly quantify their influence in a prediction model, we extend this idea and evaluate to what extent all protected characteristics jointly influence downstream predictions.
Additionally, we discuss orthogonalization as a potential solution for the embedding bias.
The authors of \cite{marcinkevics2022debiasing} tackle debiasing CXR classifiers by adapting the model used for prediction.
Our approach, in contrast, is model-agnostic and can correct any neural representation.

Bias mitigation methods are separated into pre-, in-, and post-processing methods.
Post-processing methods, like our approach, focus on the trained model or predictions as the target of their correction \cite{hort2022bia}.
An example of a pre-processing method is \cite{amini2019uncovering}, which propose to increase the sampling probability of training data with underrepresented concepts based on an unsupervised latent code obtained by an autoencoder.
The meta orthogonalization of \cite{david2020debiasing} represents an in-processing method where intermediate network layers are trained using a debiasing loss to produce activations that encourage orthogonality in different concepts (e.g., gender and class labels).

The implicit usage of protected features in predictive models prompts a discussion about their role as potential biomarkers or indicators in the context of training datasets.
However, if an accurate medical prediction supported by protected information is of importance, we argue that this relationship should be modeled explicitly by including protected information as a predictor.
For this purpose, orthogonalization helps to disentangle the influence of protected features and other factors and makes the influence of race, sex, age, or other information explicit instead of concealing it within the neural representations. 
By explicitly including metadata as separate predictors, as demonstrated in semi-structured models \cite{rugamer2023semi, rugamer2023new}, the influence of protected features can be accurately quantified and considered in the decision-making process.
This approach enables a more precise and nuanced understanding of how protected features impact predictions, which can be particularly crucial for ensuring fairness and avoiding unintended biases in medical applications.
Future work could explore how the application of orthogonalization in such a semi-structured model influences and alters saliency maps for explaining the decisions of DL models.

In practical settings, new test observations might be scarce, or information about the protected features might not be available at all. 
While our method will, by construction, even work (i.e., yield effects of 0 and p-values of 1 in a subsequent evaluation step) in the case of only a single test example, the estimated and subsequently subtracted protected feature effect(s) will have a large variance for small sample sizes.
Given that the foundation of our orthogonalization technique is rooted in linear model theory, power analysis or rule of thumb techniques such as the ``one in ten'' rule suggesting 10 samples per protected feature (c.f. \cite{freedman1983note}, or more conservative alternatives such as 1:20), analogously apply for our procedure.
In situations where information on protected attributes is absent, the corrections of embeddings themselves are not straightforward.
One alternative way to protect the information in this case, is to correct the downstream model that uses the embedding.
This can be achieved by employing training data that includes protected feature information and orthogonalizing the model’s weights as described in \cite{rugamer2023new}.
The resulting model then only requires the image information for predicting test observations.
Moreover, since the model itself is corrected to exclude the influence of protected attributes, the adjusted model weights rely solely on information orthogonal to the protected features, thus ensuring corrected predictions. 

\paragraph{Limitations.}
Our study has several limitations.
The downstream classifier is reduced to a specific model and parameter setting, which yields a fair comparison between datasets and embeddings.
We did not conduct a hyperparameter search that might enable better performance metrics; instead, our study focused on the influence of protected features concealed in a deep representation and the removal thereof.
Our analysis is further based on a linear evaluation protocol that allows the application of the orthogonalization method.
An extension of this method to incorporate non-linearities is non-trivial and is subject to future work.
Lastly, even with the orthogonalized embedding, subgroup disparities still exist, implying more complex relationships between protected features and outcomes.

\paragraph{Conclusion.}
To conclude, our study confirms the results of previous research \cite{glocker2022risk, glocker2023algorithmic}, which addresses the existence of biases in deep feature embeddings of CXR images.
We demonstrated orthogonalization as a post-hoc approach to isolate and remove protected information from neural representations and showed that orthogonalized embeddings do not allow the prediction of age, sex, and race.

\section*{Acknowledgments}

The authors gratefully acknowledge LMU Klinikum for providing computing resources on their
Clinical Open Research Engine (CORE). This work has been partially funded by the Deutsche
Forschungsgemeinschaft (DFG, German Research Foundation) as part of BERD@NFDI - grant
number 460037581.

\bibliographystyle{elsarticle-num} 
\bibliography{bibliography}

\clearpage
\appendix

\setcounter{figure}{0}    
\setcounter{table}{0}    

\renewcommand\thefigure{S.\arabic{figure}}
\renewcommand\thetable{S.\arabic{table}}

\section{Influence of Protected Features on Model Prediction}
\label{app:influence}
    
    The following figures are an extension of the results presented in Section~\ref{sec:influence} and visualize Table~\ref{tab:pred-coefs}. This includes the distribution of coefficients and p-values over ten randomly initialized runs obtained from the \textit{evaluation model}.
    The results are shown for the three exemplary labels \textit{Pleural Effusion}, \textit{Cardiomegaly}, and \textit{No Finding}.
\subsection{Pathology: Pleural Effusion}

\begin{figure}[H]
    \centering
    \begin{subfigure}{0.8\textwidth}
        \centering
        \includegraphics[width=\linewidth]{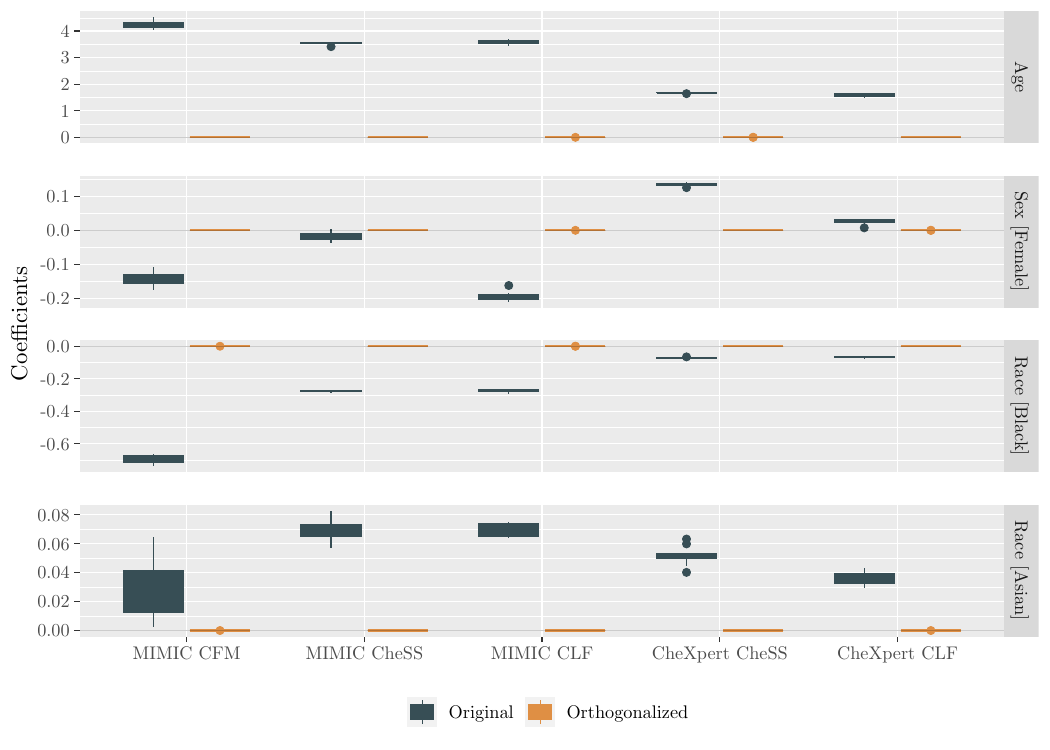}
        \caption{Coefficients for \textit{Pleural Effusion}.}
        \label{fig:influence-pleural-coef}
    \end{subfigure}
    \begin{subfigure}{0.8\textwidth}
        \centering
        \includegraphics[width=\linewidth]{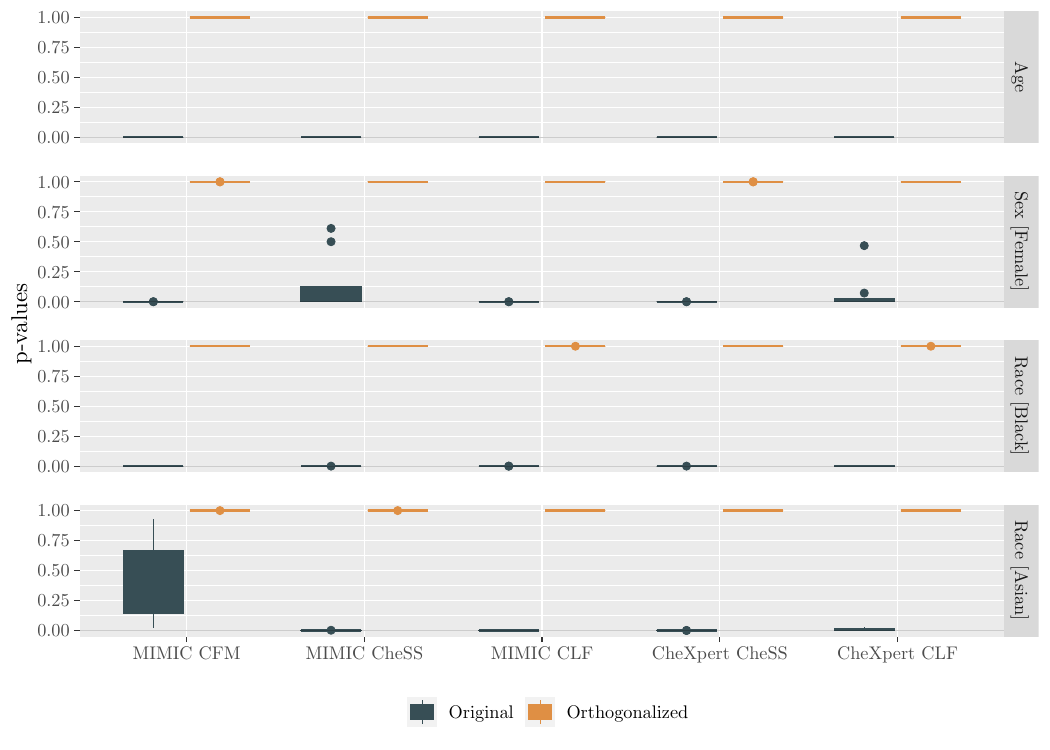}
        \caption{p-values associated with the respective coefficients.}
        \label{fig:influence-pleural-pval}
    \end{subfigure}
    \caption{Distribution of derived coefficients and p-values for 10 downstream models per embedding and protected feature category on the label \textit{Pleural Effusion}.}
    \label{fig:influence-pleural}
\end{figure}

\clearpage

\subsection{Pathology: Cardiomegaly}

\begin{figure}[H]
    \centering
    \begin{subfigure}{0.8\textwidth}
        \centering
        \includegraphics[width=\linewidth]{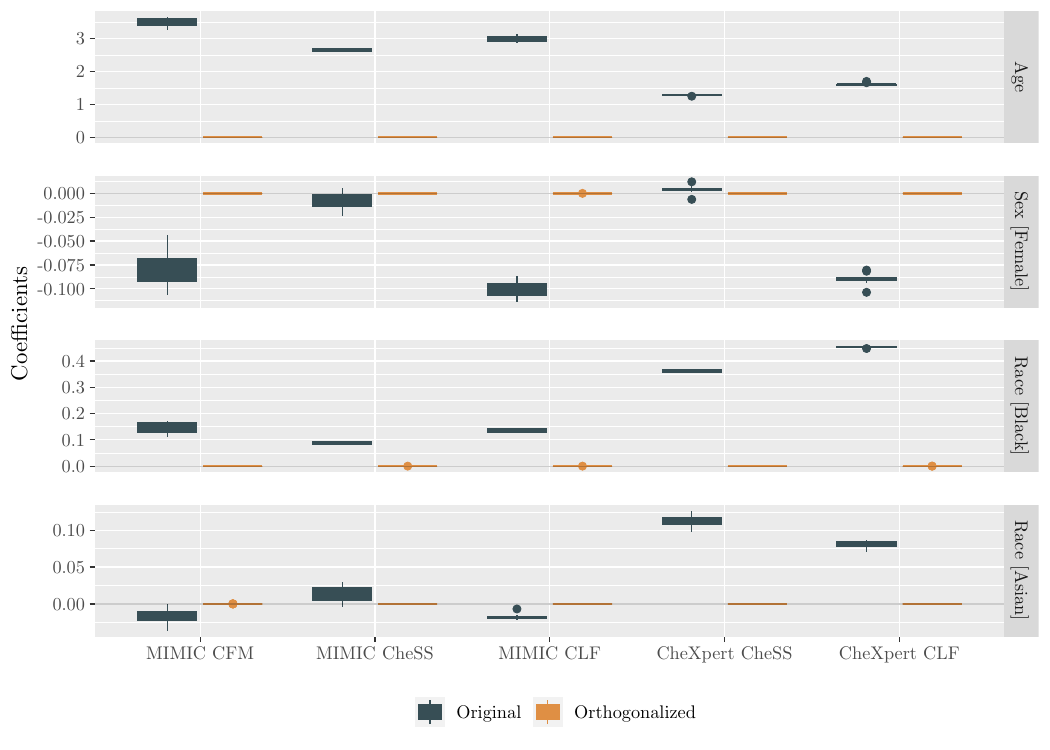}
        \caption{Coefficients for \textit{Cardiomegaly}.}
        \label{fig:influence-cardio-coef}
    \end{subfigure}
    \begin{subfigure}{0.8\textwidth}
        \centering
        \includegraphics[width=\linewidth]{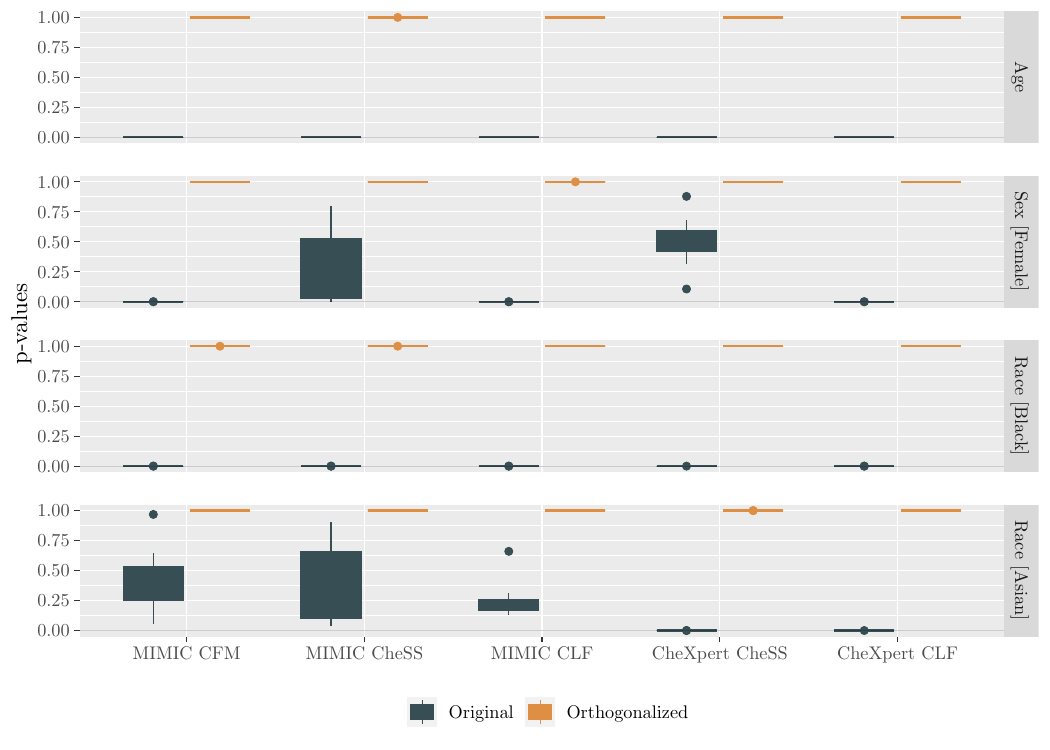}
        \caption{p-values associated with the respective coefficients.}
        \label{fig:influence-cardio-pval}
    \end{subfigure}
    \caption{Distribution of derived coefficients and p-values for 10 downstream models per embedding and protected feature category on the label \textit{Cardiomegaly}.}
    \label{fig:influence-cardio}
\end{figure}

\clearpage

\subsection{Pathology: No Finding}

\begin{figure}[H]
    \centering
    \begin{subfigure}{0.8\textwidth}
        \centering
        \includegraphics[width=\linewidth]{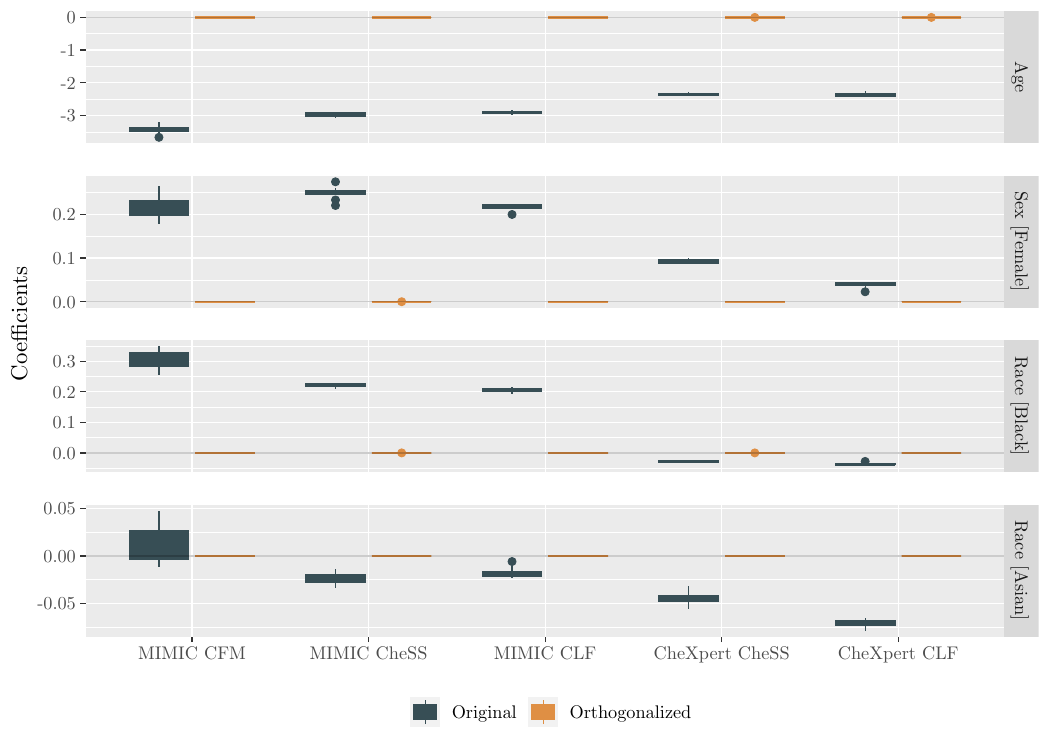}
        \caption{Coefficients for \textit{No Finding}.}
        \label{fig:influence-finding-coef}
    \end{subfigure}
    \begin{subfigure}{0.8\textwidth}
        \centering
        \includegraphics[width=\linewidth]{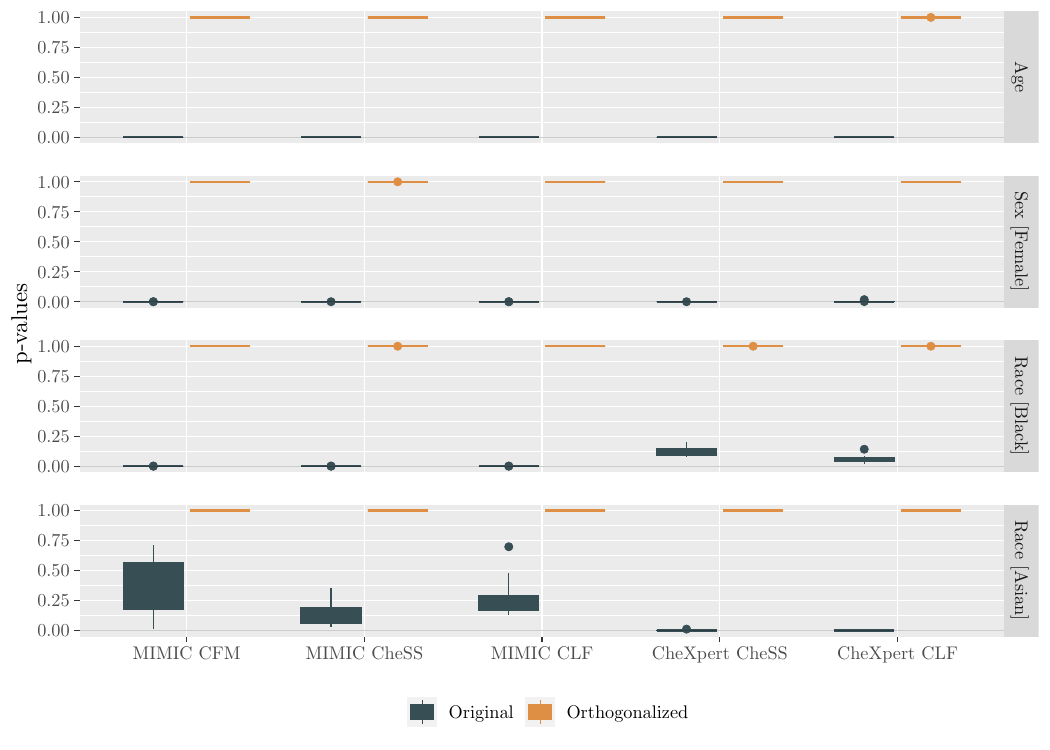}
        \caption{p-values associated with the respective coefficients.}
        \label{fig:influence-finding-pval}
    \end{subfigure}
    \caption{Distribution of derived coefficients and p-values for 10 downstream models per embedding and protected feature category on the label \textit{No Finding}.}
    \label{fig:influence-finding}
\end{figure}

\clearpage

\section{Downstream Prediction Performance}

The following table provides additional metrics for the the labels \textit{Pleural Effusion}, \textit{Cardiomegaly} and \textit{No Finding} and supplements Section~\ref{sec:res-down} and Table~\ref{tab:auc-scores}.

\begin{table}[H]
    \centering
    \begin{doublespacing}
    \resizebox{0.9\textwidth}{!}{%
    \begin{tabular}{lccc|cc|cc}
    \toprule
        \multicolumn{2}{r}{\textbf{Pathology:}}  &  \multicolumn{2}{c}{\textbf{Pleural Effusion}} & \multicolumn{2}{c}{\textbf{Cardiomegaly}} & \multicolumn{2}{c}{\textbf{No Finding}} \\
        \cmidrule{3-8}
           \multicolumn{2}{r}{\textbf{Ortho.:}} & \xmark & \cmark & \xmark & \cmark & \xmark & \cmark \\
    \midrule
\multirow{6}{*}{\rotatebox[origin=c]{90}{\textbf{MIMIC CFM}}} & AUC & 0.870 $\pm$ 0.001 & 0.856 $\pm$ 0.002 & 0.780 $\pm$ 0.001 & 0.767 $\pm$ 0.003 & 0.801 $\pm$ 0.003 & 0.786 $\pm$ 0.005 \\
& Acc. & 0.804 $\pm$ 0.003 & 0.784 $\pm$ 0.003 & 0.753 $\pm$ 0.004 & 0.751 $\pm$ 0.002 & 0.839 $\pm$ 0.005 & 0.821 $\pm$ 0.004 \\
& Sens. & 0.612 $\pm$ 0.037 & 0.485 $\pm$ 0.011 & 0.351 $\pm$ 0.083 & 0.169 $\pm$ 0.015 & 0.318 $\pm$ 0.063 & 0.474 $\pm$ 0.018 \\
& Spec. & 0.897 $\pm$ 0.016 & 0.929 $\pm$ 0.003 & 0.897 $\pm$ 0.034 & 0.961 $\pm$ 0.005 & 0.953 $\pm$ 0.018 & 0.896 $\pm$ 0.008 \\
& Prec. & 0.744 $\pm$ 0.020 & 0.768 $\pm$ 0.007 & 0.559 $\pm$ 0.029 & 0.608 $\pm$ 0.017 & 0.613 $\pm$ 0.057 & 0.500 $\pm$ 0.011 \\
& F1 & 0.670 $\pm$ 0.016 & 0.594 $\pm$ 0.008 & 0.423 $\pm$ 0.059 & 0.264 $\pm$ 0.019 & 0.411 $\pm$ 0.047 & 0.486 $\pm$ 0.008 \\
    \midrule
\multirow{6}{*}{\rotatebox[origin=c]{90}{\textbf{MIMIC CheSS}}} & AUC & 0.802 $\pm$ 0.000 & 0.792 $\pm$ 0.001 & 0.750 $\pm$ 0.000 & 0.742 $\pm$ 0.001 & 0.747 $\pm$ 0.001 & 0.737 $\pm$ 0.001 \\
& Acc. & 0.755 $\pm$ 0.002 & 0.737 $\pm$ 0.001 & 0.742 $\pm$ 0.002 & 0.740 $\pm$ 0.001 & 0.816 $\pm$ 0.004 & 0.797 $\pm$ 0.001 \\
& Sens. & 0.506 $\pm$ 0.046 & 0.320 $\pm$ 0.007 & 0.139 $\pm$ 0.048 & 0.054 $\pm$ 0.009 & 0.245 $\pm$ 0.023 & 0.372 $\pm$ 0.007 \\
& Spec. & 0.876 $\pm$ 0.022 & 0.939 $\pm$ 0.002 & 0.960 $\pm$ 0.015 & 0.987 $\pm$ 0.002 & 0.941 $\pm$ 0.010 & 0.889 $\pm$ 0.002 \\
& Prec. & 0.666 $\pm$ 0.019 & 0.716 $\pm$ 0.005 & 0.556 $\pm$ 0.011 & 0.597 $\pm$ 0.012 & 0.476 $\pm$ 0.019 & 0.423 $\pm$ 0.003 \\
& F1 & 0.573 $\pm$ 0.023 & 0.442 $\pm$ 0.006 & 0.218 $\pm$ 0.061 & 0.099 $\pm$ 0.015 & 0.322 $\pm$ 0.016 & 0.396 $\pm$ 0.005 \\

    \midrule
\multirow{6}{*}{\rotatebox[origin=c]{90}{\textbf{MIMIC CLF}}} & AUC & 0.797 $\pm$ 0.001 & 0.780 $\pm$ 0.000 & 0.737 $\pm$ 0.000 & 0.727 $\pm$ 0.001 & 0.747 $\pm$ 0.000 & 0.736 $\pm$ 0.001 \\
& Acc. & 0.748 $\pm$ 0.001 & 0.724 $\pm$ 0.002 & 0.739 $\pm$ 0.001 & 0.737 $\pm$ 0.000 & 0.819 $\pm$ 0.004 & 0.801 $\pm$ 0.003 \\
& Sens. & 0.508 $\pm$ 0.026 & 0.299 $\pm$ 0.012 & 0.065 $\pm$ 0.019 & 0.011 $\pm$ 0.002 & 0.322 $\pm$ 0.014 & 0.421 $\pm$ 0.013 \\
& Spec. & 0.864 $\pm$ 0.014 & 0.930 $\pm$ 0.005 & 0.982 $\pm$ 0.006 & 0.998 $\pm$ 0.000 & 0.928 $\pm$ 0.007 & 0.884 $\pm$ 0.007 \\
& Prec. & 0.644 $\pm$ 0.012 & 0.674 $\pm$ 0.008 & 0.573 $\pm$ 0.022 & 0.647 $\pm$ 0.063 & 0.496 $\pm$ 0.016 & 0.443 $\pm$ 0.008 \\
& F1 & 0.567 $\pm$ 0.012 & 0.414 $\pm$ 0.011 & 0.115 $\pm$ 0.031 & 0.021 $\pm$ 0.003 & 0.390 $\pm$ 0.006 & 0.431 $\pm$ 0.006 \\

    \midrule
\multirow{6}{*}{\rotatebox[origin=c]{90}{\textbf{CheXpert CheSS}}} & AUC & 0.793 $\pm$ 0.000 & 0.798 $\pm$ 0.000 & 0.789 $\pm$ 0.000 & 0.794 $\pm$ 0.001 & 0.833 $\pm$ 0.000 & 0.825 $\pm$ 0.001 \\
& Acc. & 0.726 $\pm$ 0.001 & 0.730 $\pm$ 0.000 & 0.875 $\pm$ 0.000 & 0.876 $\pm$ 0.000 & 0.915 $\pm$ 0.000 & 0.915 $\pm$ 0.000 \\
& Sens. & 0.616 $\pm$ 0.028 & 0.606 $\pm$ 0.004 & 0.053 $\pm$ 0.012 & 0.077 $\pm$ 0.006 & 0.099 $\pm$ 0.014 & 0.064 $\pm$ 0.010 \\
& Spec. & 0.800 $\pm$ 0.018 & 0.814 $\pm$ 0.003 & 0.996 $\pm$ 0.001 & 0.994 $\pm$ 0.001 & 0.992 $\pm$ 0.001 & 0.995 $\pm$ 0.001 \\
& Prec. & 0.679 $\pm$ 0.010 & 0.690 $\pm$ 0.002 & 0.642 $\pm$ 0.027 & 0.640 $\pm$ 0.019 & 0.546 $\pm$ 0.012 & 0.564 $\pm$ 0.016 \\
& F1 & 0.645 $\pm$ 0.011 & 0.645 $\pm$ 0.002 & 0.097 $\pm$ 0.020 & 0.138 $\pm$ 0.009 & 0.167 $\pm$ 0.020 & 0.115 $\pm$ 0.015 \\
    \midrule
\multirow{6}{*}{\rotatebox[origin=c]{90}{\textbf{CheXpert CLF}}} & AUC & 0.804 $\pm$ 0.000 & 0.801 $\pm$ 0.000 & 0.799 $\pm$ 0.000 & 0.794 $\pm$ 0.001 & 0.854 $\pm$ 0.000 & 0.844 $\pm$ 0.000 \\
& Acc. & 0.732 $\pm$ 0.001 & 0.731 $\pm$ 0.001 & 0.878 $\pm$ 0.000 & 0.878 $\pm$ 0.000 & 0.916 $\pm$ 0.000 & 0.915 $\pm$ 0.000 \\
& Sens. & 0.686 $\pm$ 0.021 & 0.670 $\pm$ 0.006 & 0.102 $\pm$ 0.014 & 0.122 $\pm$ 0.007 & 0.144 $\pm$ 0.029 & 0.161 $\pm$ 0.017 \\
& Spec. & 0.764 $\pm$ 0.014 & 0.772 $\pm$ 0.003 & 0.991 $\pm$ 0.002 & 0.989 $\pm$ 0.001 & 0.989 $\pm$ 0.003 & 0.987 $\pm$ 0.002 \\
& Prec. & 0.665 $\pm$ 0.006 & 0.668 $\pm$ 0.001 & 0.634 $\pm$ 0.021 & 0.619 $\pm$ 0.007 & 0.563 $\pm$ 0.013 & 0.541 $\pm$ 0.010 \\
& F1 & 0.675 $\pm$ 0.007 & 0.669 $\pm$ 0.003 & 0.175 $\pm$ 0.020 & 0.204 $\pm$ 0.009 & 0.228 $\pm$ 0.034 & 0.248 $\pm$ 0.018 \\

    \end{tabular}
    }
    \end{doublespacing}
    \caption{Prediction performance original versus orthogonalized data on the MIMIC and CheXpert datasets. The table shows the mean and standard deviation over 10 randomly initialized runs for the labels \textit{Pleural Effusion}, \textit{Cardiomegaly} and \textit{No Finding} and the metrics AUC, accuracy, sensitivity, specificity, precision, and F1-score.}
    \label{tab:pred-scores}
\end{table}

\end{document}